\documentclass[10pt,twocolumn,letterpaper]{article}

\usepackage{iccv}
\usepackage{times}
\usepackage{epsfig}
\usepackage{graphicx}
\usepackage{amsmath}
\usepackage{amssymb}

\usepackage{bbm}
\usepackage{subfigure}

\usepackage{colortbl}
\usepackage{xcolor}
\usepackage{array}
\usepackage{enumitem}
\setlist[itemize]{noitemsep, nolistsep}
\definecolor{mygray}{gray}{.9}

\usepackage[accsupp]{axessibility}  

\usepackage[pagebackref=true,breaklinks=true,letterpaper=true,colorlinks,bookmarks=false]{hyperref}

\iccvfinalcopy 

\def\iccvPaperID{7373} 

\ificcvfinal\fi

\begin{document}


\title{DomainDrop: Suppressing Domain-Sensitive Channels for \\Domain Generalization}



\author{
Jintao~Guo$^{1,2}$ \;\; Lei Qi$^{3,*}$ \;\; Yinghuan Shi$^{1,2,*}$\\
$^1$~State Key Laboratory for Novel Software Technology, Nanjing University \\
$^2$~National Institute of Healthcare Data Science, Nanjing University \\
$^3$~School of Computer Science and Engineering, Southeast University \\
{\tt\small guojintao@smail.nju.edu.cn, qilei@seu.edu.cn, syh@nju.edu.cn}
}

\maketitle
\ificcvfinal\thispagestyle{empty}\fi

\renewcommand{\thefootnote}{\fnsymbol{footnote}}
\footnotetext[1]{Corresponding authors: Yinghuan Shi and Lei Qi. 
The work is supported by NSFC Program (62222604, 62206052, 62192783), China Postdoctoral Science Foundation Project (2023T160100), Jiangsu Natural Science Foundation Project (BK20210224), and CCF-Lenovo Bule Ocean Research Fund.
}

\begin{abstract}
Deep Neural Networks have exhibited considerable success in various visual tasks. However, when applied to unseen test datasets, state-of-the-art models often suffer performance degradation due to domain shifts. In this paper, we introduce a novel approach for domain generalization from a novel perspective of enhancing the robustness of channels in feature maps to domain shifts. We observe that models trained on source domains contain a substantial number of channels that exhibit unstable activations across different domains, which are inclined to capture domain-specific features and behave abnormally when exposed to unseen target domains. To address the issue, we propose a DomainDrop framework to continuously enhance the channel robustness to domain shifts, where a domain discriminator is used to identify and drop unstable channels in feature maps of each network layer during forward propagation. We theoretically prove that our framework could effectively lower the generalization bound. Extensive experiments on several benchmarks indicate that our framework achieves state-of-the-art performance compared to other competing methods. Our code is available at \textcolor{magenta}{\href{https://github.com/lingeringlight/DomainDrop}{https://github.com/lingeringlight/DomainDrop}}.
\end{abstract}



\section{Introduction}

Deep neural networks (DNNs) have shown impressive performance in computer vision tasks over the past few years. However, this performance often degrades when the test data follows a different distribution from the training data \cite{li2017deeper}. This issue, known as domain shift \cite{pan2009survey}, has greatly impaired the applications of DNNs \cite{mahajan2021domain,xu2021fourier}, as training and test data often come from different distributions in reality. To address this issue, domain adaptation (DA) has been widely studied under the assumption that some labeled or unlabeled target domain data can be observed during training \cite{du2021cross,wei2021metaalign}. Despite their success, DA models cannot guarantee their performance on unknown target domains that have not been seen during training, which makes them unsuitable for some real-world scenarios where target data are not always available \cite{wang2021learning}. Therefore, domain generalization (DG) is proposed as a more challenging yet practical setting, which aims to utilize multiple different but related source domains to train a model that is expected to generalize well on arbitrary unseen target domains \cite{zhou2021survey,wang2022generalizing}.

  \begin{figure}[tbp!]
    \centering
    \includegraphics[width=0.95\linewidth]{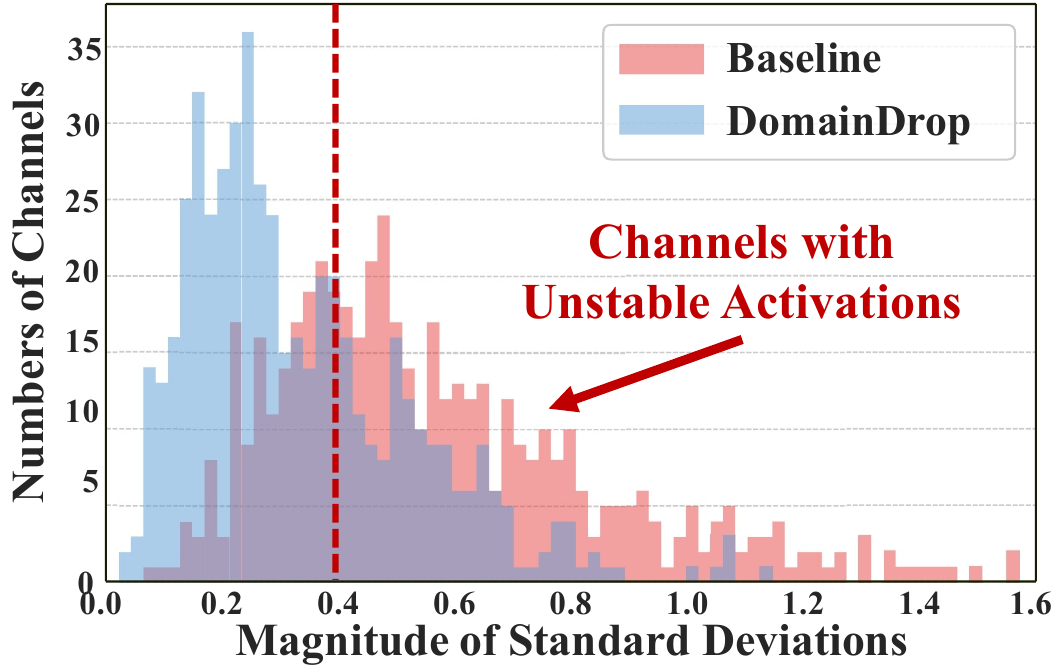}
    \centering
    \label{fig:channel sensitivity}
    \vspace{0.1cm}
    \caption{
        The robustness of channel to domain shifts.
        We investigate channel robustness using the histogram of activations based on their standard deviation across different domains. 
        We experiment on PACS \cite{li2017deeper} with sketch as the target domain and analyze the representations from the last residual block of ResNet-$18$.
        For each channel, averaged activations are computed across all samples from each domain, and the standard deviation is calculated on domain dimension to indicate its robustness to domain shifts.
        }
    \label{fig:layer domain gap}
    \vspace{-0.1cm}
  \end{figure}

The core idea of existing DG methods is to learn domain-invariant feature distribution $P(F(X))$ across domains for the robustness of conditional distribution $P(Y|F(X))$, where $F(X)$ denotes the extracted features from input $X$, and $Y$ is the corresponding label. Traditional DG methods primarily impose constraints on the whole network (\ie, the prediction layer) to supervise the model to learn domain-invariant features \cite{zhu2022localized,lee2022cross}. However, these methods do not explicitly guide the model to remove domain-specific features in middle network layers, which could lead to the model learning excessive domain-related information. In this paper, we revisit DG issue from a novel perspective of feature channels, which indicates that model generalization could be related to the robustness of feature channels to domain shifts.
Specifically, we quantify the robustness of each channel to domain shifts by computing the standard deviation of activations across different domains. As shown in Fig.~\ref{fig:channel sensitivity}, we observe that models trained on source domains often contain numerous non-robust channels that exhibit unstable activations for different domains, indicating that they are likely to capture domain-specific features. When domain shifts, these unstable channels are likely to produce abnormal activations on the unseen target domain, leading to a shift in the conditional distribution. These unstable channels are dubbed as ``domain-sensitive channels".

Based on the above observation, we propose \textit{DomainDrop}, a simple yet effective framework that explicitly mutes unstable channels across domains. Unlike previous DG methods that seek to directly distill domain-invariant features, our method aims to continuously guide the model to remove domain-specific features during forward propagation. To this end, we introduce a domain discriminator to assign each channel a specific dropout rate according to its effectiveness for the domain discrimination task. The more the channel contributes to domain prediction, the more likely it contains domain-specific features, and the greater the probability of it being discarded. Moreover, we discover that unstable channels exist at both shallow and deep network layers, which contrasts with existing dropout methods that drop either high-level or low-level features. Thus, we propose a \textit{layer-wise training strategy} that inserts DomainDrop at a random middle layer of the network at each iteration, which can sufficiently narrow domain gaps in multiple network layers. 
To further enhance the robustness of channels against domain shifts, we adopt a \textit{dual consistency loss} to regularize the model outputs under various perturbations of DomainDrop. 
Furthermore, we provide theoretical evidence that our method could effectively lower the generalization error bound of the model on unseen target domains. 


Our contributions can be summarized as follows:

\begin{itemize}[itemsep=5pt,topsep=4pt]
    \item We propose a novel dropout-based framework for DG, which explicitly suppresses domain-sensitive channels to enhance the robustness of channels to domain shifts.
    \item We theoretically prove that removing domain-sensitive channels during training could result in a tighter generalization error bound and better generalizability.
    \item We evaluate our method on four standard datasets. The results demonstrate that our framework achieves state-of-the-art performance on all benchmarks.
\end{itemize}

\section{Related Works}

\textbf{Domain generalization.}
Domain generalization (DG) aims to extract knowledge from source domains that is well-generalizable to unseen target domains.
One prevalent approach is to align the distributions of source domains by \textit{learning domain-invariant representations} via 
adversarial learning \cite{zhu2022localized}, causality learning \cite{sun2021recovering,lv2022causality} or meta-learning methods \cite{zhang2022mvdg,wei2021metaalign}.
Another important method is \textit{domain augmentation}, which 
empowers the model with generalization ability by enriching the diversity of source data at image-level \cite{zhou2020deep,cubuk2020randaugment} or feature-level \cite{wang2022feature,li2021simple}. 
Although demonstrating promising results, 
these methods may still learn excessive domain-specific features, as they rely on the hope that domain-specific features would be implicitly removed by achieving the final goal of learning domain-invariant features via image-level augmentations or model-level constraints.
Recently, some works reveal that CNNs tend to classify objects based on features from superficial local textures that are likely to contain domain-specific features \cite{wang2019learning}. They propose to penalize the model from learning local representation and make the CNNs rely on global representations for classification \cite{wang2019learning,shi2020informative}. 
However, the local features may be only one kind of domain-specific feature, and there could exist other forms of domain-specific features that lead to the overfitting issue. 
Different from these methods, our framework is proposed to continuously drop domain-sensitive channels during forward propagation, which could explicitly suppress the model learning of generic domain-specific features.

\vspace{0.1cm}
 \textbf{Dropout regularization.} 
Since our method builds on discarding domain-specific features during training, we here compare our method with dropout-based methods. 
As one of the most widely used regularization methods, dropout \cite{srivastava2014dropout} aims to fight the overfitting issue by randomly discarding neurons.
SpatialDropout \cite{tompson2015efficient} is proposed to randomly drop features across channels that capture different patterns. 
DropBlock \cite{ghiasi2018dropblock} is designed to mask random contiguous regions within a feature map.
Recently, a series of structure-information dropout methods have also emerged, which utilize feature-level structure information to guide dropout operations \cite{hou2019weighted,zeng2021correlation,guo2021domain}. 
DAT \cite{lee2019drop} uses adversarial dropout based on cluster assumption to help the model learn discriminative features.
RSC \cite{huang2020self} attempts to mute the most predictive parts of feature maps for learning comprehensive information. 
PLACE \cite{guo2021domain} seeks to activate diverse channels by randomly dropping feature channels for mitigating model overfitting.
However, these methods may not be effective in the DG task where there exists a large distribution gap between source and unseen target domains \cite{ding2022domain,du2022cross}. 
Due to the lack of guidance on suppressing domain-specific information, these methods still inevitably learn excessive domain-specific features and suffer from the overfitting issue on source domains. 
As a test-time adaptation method, SWR \cite{choi2022improving} is proposed to enable rapid adaptation to target domains during test, which reduces the update of shift-agnostic parameters identified by specific transformations, but could not address other sensitive parameters unaffected by the transformations.
To address these problems, we explore a novel domain discriminator guided dropout for DG, which explicitly identifies and discards unstable channels that are non-robust to domain shifts using domain discriminators. 
The proposed method works only during training, relying on no target data or specific transformations. 
Besides, different from the most related works that solely employ dropout on either low-level \cite{du2022cross,shi2020informative} or high-level features \cite{huang2020self,lv2022causality}, 
our method involves muting domain-specific information across all network layers, which effectively combats the overfitting issue of the model.

\begin{figure*}[tb!]
    \begin{center}
    \includegraphics[width=0.95\linewidth]{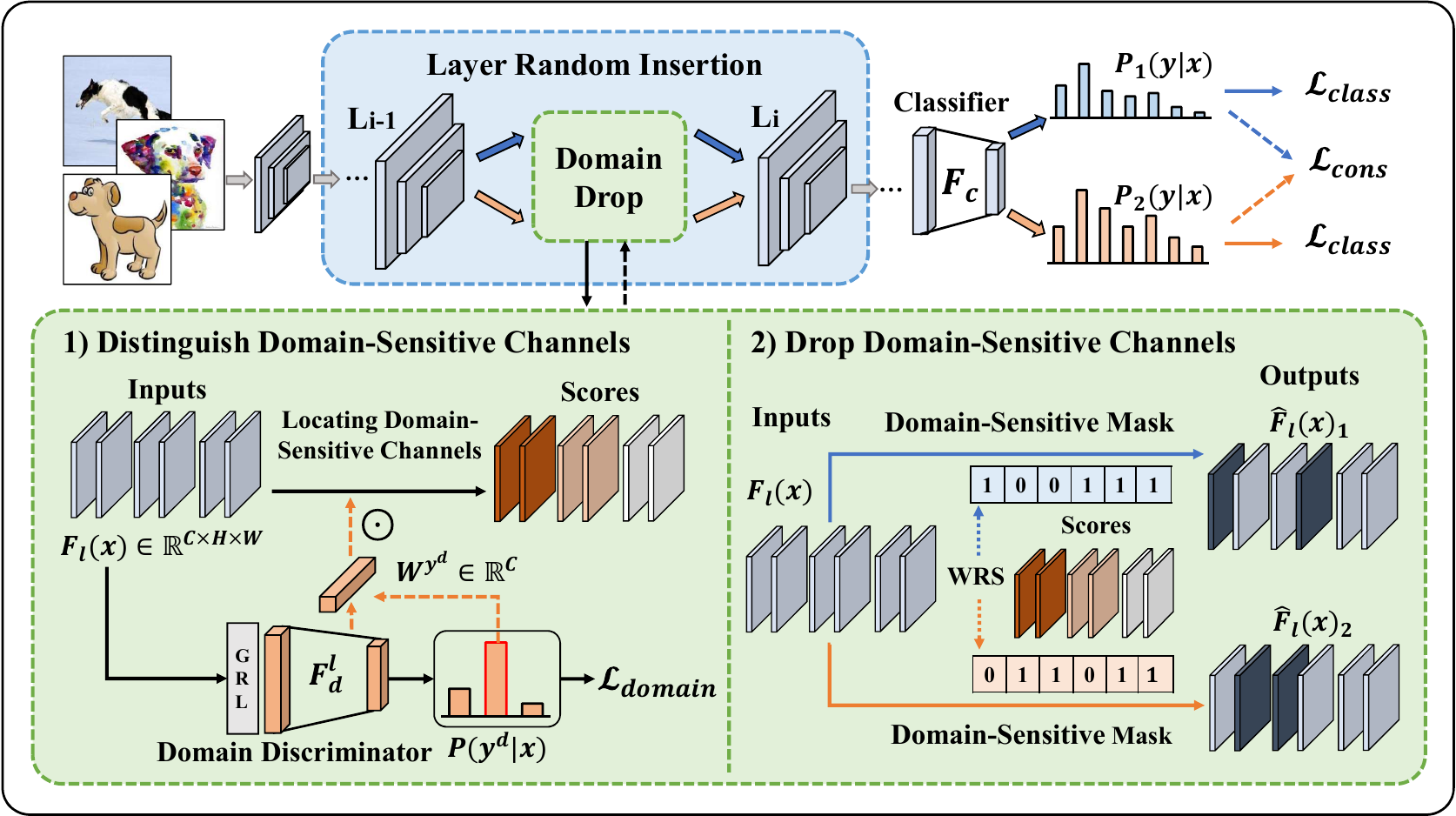}
    \end{center}
    \vspace{-0.1cm}
    \caption{
        An overview of the proposed framework. Our framework contains three key components, including the DomainDrop, the layer-wise training scheme, and the dual consistency loss. At each iteration, we randomly select one middle layer to apply DomainDrop, which uses a domain discriminator to locate and drop domain-sensitive channels. 
        To further enhance channel stability to domain shifts, we utilize the dual consistency loss that aligns the model predictions for the same sample under different perturbations generated by DomainDrop.
    }
    \label{fig:DomainDrop framework}
    \vspace{-0.2cm}
\end{figure*}

\section{Methodology}

\subsection{Setting and Overview}
Assuming that we are given $K$ observed source domains $\mathcal{D}_s = \{D_s^1, D_s^2, ..., D_s^K\}$ that follow different disrtibutions. For each domain, $D_s^k = \{(x_i^k, y_i^k)\}_{i=1}^{n_k}$, where $n_k$ is the number of samples in $D_s^k$, and $(x_i^k, y_i^k)$ denotes the sample-label pair for the $i$-th sample in the $k$-th domain. 
The goal of domain generalization is to use the source domains $\mathcal{D}_s$ to train a model $F$ that is expected to perform well on unseen target domain $\mathcal{D}_t$.
In general, the model $F$ can be divided into a feature extractor $F_f$ and a classifier $F_c$, respectively. 

With training data of source domains, our framework trains the model with the standard classification loss.
However, unlike previous DG methods, we aim to explicitly remove domain-specific features while enhancing domain-invariant ones. 
To this end, we propose a novel technique, called DomainDrop, which uses a domain discriminator to distinguish and remove domain-sensitive channels. 
Besides, we observe that domain-sensitive channels exist in each network layer, thus designing a layer-wise training scheme that applies DomainDrop to both high-level and low-level layers. 
Moreover, we add a dual consistency loss to reach consensuses between predictions derived by the model under different perturbations of DomainDrop, which could further reduce channel instability to domain shifts and enhance the model learning of domain-invariant features.
The overview of our framework is illustrated in Fig.~\ref{fig:DomainDrop framework}. 
Below we introduce the main components of our framework and provide a theoretical analysis of its effectiveness.

\subsection{Suppressing Domain-Sensitive Channels}
To clearly inform the model to remove domain-specific features during training, we introduce domain discriminators to multiple middle layers for locating domain-sensitive channels, which consists of a Global Average Pooling (GAP) layer and a Fully-Connected (FC) layer.
Given an input $x_i^k$ from source domain $D_k$ and its label $y_i^k$, we first extract the feature $F_l(x_i^k) \in \mathbb{R}^{C \times H \times W}$ that is yielded by the $l$-th middle layer, where $C$ is the number of channels, $H$ and $W$ denote the height and width dimensions, respectively.
The feature map $F_l(x_i^k)$ is fed to domain discriminator $F^l_d$ to predict domain labels and compute discrimination loss. 
To avoid the negative impact of domain discriminators on the main network 
, we use a gradient reversal layer (GRL) \cite{matsuura2020domain} before the domain discriminator to truncate the gradients of minimizing discrimination loss: 
\begin{equation}
    \mathcal{L}^l_{domain} = -\frac{1}{K} \sum_{k=1}^{K} \left[\frac{1}{n_k} \sum_{i=1}^{n_k} \sum_{j=1}^{K} \mathbbm{1}_{[j=k]} \log F_d(F_l(x_i^k))\right].
    \label{eq:domain loss}
    \vspace{0.1cm}
\end{equation}


\textbf{Distinguish domain-sensitive channels.}
To determine which channels contain domain-specific information, we use the performance of the domain discriminator in the middle layer as an indicator of channel importance. 
Specifically, we hypothesize that \textit{channels that contribute the most to domain prediction are likely to contain domain-specific information}.
We quantify the correlation between each channel and domain-specific information by computing the weighted activations for the correct domain prediction. 
For an input $x_i^k$ and the feature extractor $F_l(\cdot)$, we define the score of the $j$-th channel in the feature map $F_l(x_i^k)$ as:
\vspace{0.05cm}
\begin{equation}
    s_j = W^{y^d}_j \cdot \text{GAP}(F_l(x_i^k))_j,
    \label{eq:channel score}
    \vspace{0.05cm}
\end{equation}
where $W^{y^d} \in \mathbb{R}^C$ is the FC layer weight of the domain discriminator $F_d(\cdot)$ for the true domain $y^d$, and $C$ is the channel number.
The higher the weighted activation value, the more contribution of the channel to the true domain prediction. 
Thus, we aim to reduce the impact of domain-sensitive channels on the classification task by constraining domain-specific information used by the domain discriminator.


\textbf{Dropping domain-sensitive channels.} To explicitly reduce domain-specific information in the feature maps, we propose to select and drop the most domain-sensitive channels during training.
Specifically, with score $s_j$, we first calculate the probability $p_j$ of discarding the $j$-th channel:
\vspace{-0.1cm}
\begin{equation}
    p_j = s_j / \sum_{c=1}^{C} s_c.
    \label{eq:channel prob}
    \vspace{-0.1cm}
\end{equation}
Subsequently, we generate a binary mask $m \in \mathbb{R}^{C}$ based on the probability of each channel, \ie, $m_j$ has the probability $p_j$ of being set to $1$, with channels with relatively high $s_j$ more likely to be discarded. 
To ensure the regularization effect of dropout, we attempt to drop a certain number of domain-specific channels by probability $p_j$.
The naive algorithm to achieve this has been analyzed in \cite{efraimidis2006weighted}, but it suffers a high time complexity of $O(C \times M)$, where $C$ is the number of channels and $M$ is the number of discarded channels.
To reduce computation cost, we employ weighted random selection (WRS) algorithm \cite{efraimidis2006weighted,hou2019weighted} to generate the binary mask $m$, which enjoys the time complexity as $O(C)$. Specifically, for the $j$-th channel with $s_j$, we first generate a random number $r_j \in (0, 1)$ and compute a key value $k_j = r_j^{1/s_j}$. 
Then we select $M$ items with the largest key values and set the corresponding $m_j$ of the mask $m$ to $0$:
\begin{equation}
    m_{j} = 
    \begin{cases}
    0,& \text{if}\enspace j \in \text{TOP}(\{k_1, k_2, ..., k_C\}, M) \\
    1,& \text{otherwise}
    \end{cases},
    \label{eq:channel mask}
\end{equation}
where $j$ is the channel index, TOP($\{k_1, k_2, ..., k_C\}$, $M$) denotes the $M$ items with the largest key value $k$.
Additionally, the hyper-parameter $P_{drop} = \frac{M}{C}$ indicates the number of channels to discard.
In practice, we use a hyperparameter $P_{active}$ to control the activation probability of DomainDrop in the forward pass, which can narrow the domain gap by preserving the original feature maps while meantime reducing domain-specific information.
During inference, DomainDrop is closed as conventional dropout \cite{srivastava2014dropout}.

\textbf{Remark.} 
Note that DomainDrop differs significantly from previous DG methods that focus on constraining the network to extract domain-invariant information during the backpropagation stage.
In contrast, our DomainDrop operates during the feature forward propagation stage, which continuously filters out domain-sensitive channels to prevent the model from retaining too many domain-related features.
The process could also be regarded as a guided feature-level augmentation that effectively disturbs domain-specific features while preserving domain-invariant features.
Compared to previous dropout methods, we address the DG issue from a novel perspective, considering enhancing channel robustness to domain shifts. 
By dropping generic domain-sensitive channels during training, our method reduces channel instability caused by domain shifts and promotes the learning of domain-invariant features.

\subsection{Layer-wise Training Scheme}

Prior research has revealed that CNNs encode information across multiple stages, with shallow layers capturing more generic patterns and deep ones encoding more semantic features \cite{zeiler2014visualizing,zhang2021can}. 
Traditional dropout techniques focus on removing information from high-level semantic layers that contain crucial classification information \cite{lee2019drop,huang2020self,zeng2021correlation,lv2022causality}.
Recently, some studies suggest that features extracted from shallow layers exhibit more distinct domain-specific characteristics, and thus propose gating low-level information to mitigate domain gaps \cite{shi2020informative,du2022cross}.  
However, previous methods are typically designed for a specific single layer, either low-level or high-level, which may not be optimal for DG task. 
In contrast, we propose gating feature maps from all network layers, recognizing that features extracted from each layer may contain domain-specific characteristics. 
To validate the statement, an experiment is conducted on the PACS dataset, in which we insert multiple domain discriminators with GRL (\ie, truncate gradients) in each layer and train them with the backbone network.
The accuracy of each discriminator indicates how many domain-specific features are contained in the corresponding layer.

\begin{figure}[tb!]
    \begin{center}
    \includegraphics[width=0.95\linewidth]{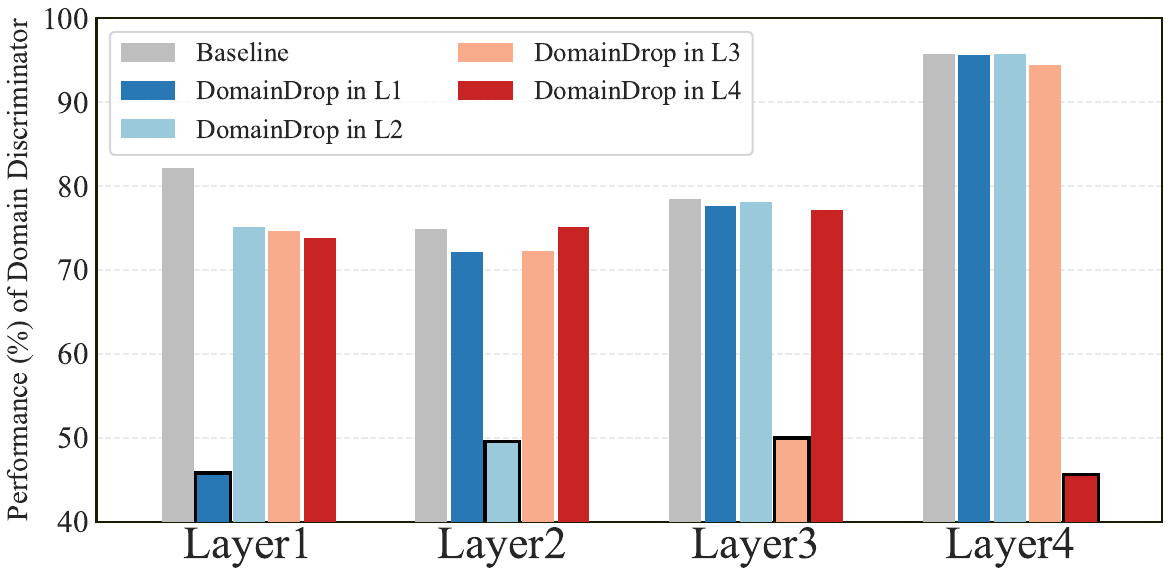}
    \end{center}
    \vspace{-0.1cm}
    \caption{The accuracy of domain discriminator in each network layer. We test the discrimination accuracies of the baseline and the models with DomainDrop in different layers. The experiment is conducted on the PACS dataset with ResNet-$18$ backbone. }
    \label{fig:layer acc}
    \vspace{-0.2cm}
\end{figure}

As shown in Fig.~\ref{fig:layer acc}, there are several observations: (1) The domain discriminator in each layer has relatively good performance, indicating that considerable domain gaps exist in every network layer. 
(2) The discrimination accuracy drops while inserting DomainDrop into the corresponding layer, suggesting that DomainDrop can effectively narrow the domain gap in the inserted layer. 
Hence, the proposed DomainDrop focuses on removing domain-specific features in both high-level and low-level layers. 
Since discarding features at multiple layers simultaneously may lead to excessive information absence and hinder model convergence \cite{park2016analysis}, we propose a layer-wise training scheme that randomly selects a middle layer of the network and performs DomainDrop on its feature maps at each iteration. 
This scheme enables DomainDrop to effectively reduce channel sensitivity to domain shifts in multiple network layers.


\vspace{0.3cm}
\subsection{Enhancing Domain-Invariant Channels}
In the training stage, DomainDrop is employed to remove domain-sensitive channels by probability, 
which can be approximated as applying Gaussian multiplicative noise to domain-specific features \cite{srivastava2014dropout,park2016analysis}. 
In light of the output inconsistency that arises from different perturbations of DomainDrop, we propose a dual consistency loss that enhances model robustness to domain-specific feature perturbations and facilitates the learning of domain-invariant features.

Specifically, for a given input data $x_i^k$, denoted as $x$ for simplicity, we apply DomainDrop twice at each training step to obtain two sets of class predictions denoted as $\widehat{F}(x)_1$ and $\widehat{F}(x)_2$, respectively. 
According to Eqs.~(\ref{eq:channel prob}) and (\ref{eq:channel mask}), the masks $m$ generated by DomainDrop could be different for the same input, leading to different predictions $\widehat{F}(x)_1$ and $\widehat{F}(x)_2$. 
To ensure consistency between these two outputs for the same input, we introduce the following constraint:
\vspace{0.1cm}
\begin{equation}
    \begin{aligned}
        \mathcal{L}_{cons} = &\frac{1}{2} (\text{KL}[\sigma(\widehat{F}(x)_1 / T) || \sigma(\widehat{F}(x)_2 / T)] \\ 
        & + \text{KL}[\sigma(\widehat{F}(x)_2 / T) || \sigma(\widehat{F}(x)_1 / T)]),
    \end{aligned}
    \label{eq:consistency}
    \vspace{0.2cm}
\end{equation}
where $\sigma$ denotes a softened softmax at a temperature $T$, and $KL$ is Kullback-Leibler divergence \cite{hinton2015distilling}. 
With this consistency constrain, our framework encourages the model to improve the robustness of channels to domain shifts
and extract domain-invariant features from perturbed representations. 
Moreover, Since DomainDrop only operates during training, the loss could also reduce the inconsistency existing in the training and inference stages \cite{wu2021r}, thus improving the generalization ability of the model to target domains.

\vspace{0.4cm}
\subsection{Theoretical Analysis of DomainDrop} 
\label{sec:theory}
Previous DG methods mainly train the model to directly distill domain-invariant features, which could still retain excessive domain-specific features and perform inferiorly when domain shifts. 
We here theoretically prove that \textit{removing the domain-sensitive channels during training can improve the generalizability in DG task}. 
Given a hypothesis $h: \mathcal{X} \rightarrow \mathcal{Y}$, where $h$ comes from the space of the candidate hypothesis $\mathcal{H}$, 
$\mathcal{X}$ and $\mathcal{Y}$ are the input space and the label space, respectively. 
Let $\phi(\cdot): \mathcal{X}\rightarrow \mathbb{R}^{n}$ be the feature extractor that maps the input images into the $n$-dimentional feature space.
Following the popular Integral Probability Metrics (IPMs) \cite{geng2011daml,zhang2012generalization}, we first define the channel-level maximum mean discrepancy (CMMD) that estimates the distribution gap between different domains by channels.

\vspace{0.3cm}
\noindent \textbf{Definition~1.} \textit{Let $n$ denote the number of channels in the extracted features of $\phi(\cdot)$. Given two different distribution of $D_s$ and $D_t$, the \textbf{channel-level maximum mean discrepancy} (CMMD) between $\phi(D_s)$ and $\phi(D_t)$ is defined as:} 
\begin{equation}
    \begin{aligned}
        d_{\text{CMMD}}(D_s, D_t)=\frac{1}{n} \sum_{i=1}^{n}\sup_{\phi_i \in \Phi_i}||\int_x &k(x, \cdot)d(\phi_i(D_s) \\
        &- \phi_i(D_t))||_{\mathcal{H}_k},
        \label{eq:CMMD paper}
    \end{aligned}
\end{equation}
\textit{
where $\Phi$ is the space of candidate hypothesis for each channel, $\phi_i(D)$ is the distribution of the $i$-th channel for the domain $D$, and $\mathcal{H}_k$ is a RKHS with its associated kernel $k$.} 

\vspace{0.3cm}
Based on the above definition, we derive the theorem that provides an upper bound on the generalization risk of the hypothesis $h$ on the target domain $D_t$.
First, we define the risk of $h$ on a domain $D$ as: $\mathcal{R}[h] = \mathbb{E}_{x \sim D} \ell[h(x), f(x)]$, where $\ell_{h,f}: x\rightarrow \ell[h(x), f(x)]$ is a convex loss-function defined for $\forall h, f \in \mathcal{H}$, and assume that $\ell$ obeys the triangle inequality. 
Following \cite{ding2022domain,zhao2022test}, for the source domains $\mathcal{D}_s = \{D_s^1, D_s^2, ..., D_s^K\}$, we define the convex hull $\Lambda_s$ as a set of mixture of source domain distributions: 
$\Lambda_s=\{\bar{D}: \bar{D}(\cdot) = \sum_{i=1}^K \pi_i D_s^i(\cdot), \pi_i \in \Delta_K\}$, where $\pi$ is non-negative coefficient in the $K$-dimensional simplex $\Delta_K$. 
We define $\bar{D}_t \in \Lambda_s$ as the closest domain to the target domain $D_t$.

\vspace{0.2cm}
\noindent \textbf{Theorem~1} \textbf{(Generalization risk bound)}. \textit{
    With the previous settings and assumptions, let $S^i$ and $T$ be two samples of size $m$ drawn i.i.d from $D_s^i$ and $D_t$, respectively.
    Then, with the probability of at least $1-\delta$ ($\delta \in (0, 1)$) for all $h \in \mathcal{F}$, the following inequality holds for the risk $\mathcal{R}_t[h]$:
}
\begin{equation}
    \begin{aligned}
        \mathcal{R}_t[h] \leq & \sum_{i=1}^{N}\pi_i\mathcal{R}_s^i[h] + d_{\text{CMMD}}(\bar{D}_t, D_t) \\
        & + \sup_{i, j \in [K]} d_{\text{CMMD}}(D_s^i, D_s^j) 
        + \lambda + \epsilon,
    \label{eq:generalization bound paper}
    \end{aligned}
\end{equation}
\textit{
    where $\lambda = 2\sqrt{\frac{\log(\frac{2}{\sigma})}{2m}} + \frac{2}{m} (\sum_{i=1}^{N}\pi_i 
    E_{x \sim D_s^i}[\sqrt{tr(K_{D_s^i})}] + E_{x\sim D_t}[\sqrt{tr(K_{D_t})}])$, $K_{D_s^i}$ and $K_{D_t}$ are kernel functions computed on samples from $D_s^i$ and $D_t$, respectively. $tr(\cdot)$ is the trace of input matrix and $\epsilon$ is the combined error of ideal hypothesis $h^*$ on $D_t$ and $\bar{D}_t$. Let $\gamma=d_{\text{\rm CMMD}}(\bar{D}_t, D_t)$ and $\beta=sup_{i, j \in [K]} d_{\text{\rm CMMD}}(D_s^i, D_s^j)$, respectively.
}

\vspace{0.1cm}
\noindent\textbf{Proof.} See in Supplementary Material. $\hfill \blacksquare$
\vspace{0.2cm}

Theorem~$1$ indicates that the upper bound of generalization risk on target domain depends mainly on: 1) $\gamma$ that measures the maximum discrepancy between different activations for the same channel in source and target domains; 2) $\beta$ that presents the maximum pairwise activation discrepancy among source domains at channel level.
Recall that our DomainDrop actively removes domain-sensitive channels during training, which can effectively restrict the size of the hypothesis space $\Phi$ and promote the model learning of domain-invariant channels.
Consequently, the features extracted by the DomainDrop model from source domains would exhibit a smaller distribution gap than the original model, \ie, effectively reducing $\beta$ in Eq.~(\ref{eq:generalization bound paper}). 
Moreover, after removing domain-sensitive channels, the features extracted from target domain would become more similar to those of source domains, thus potentially reducing $\gamma$ in Eq.~(\ref{eq:generalization bound paper}). 
Concerning Theorem $1$, \textit{by explicitly removing domain-sensitive channels, we can obtain a lower error bound and expect a better generalization ability.} 
The conclusion is also proved in the experiment section (Sec.~\ref{sec:prove theory}).

\section{Experiment}

\subsection{Datasets}
We evaluate our method on four conventional DG benchmarks: 
(1) \textbf{PACS} \cite{li2017deeper} consists of images from $4$ domains: Photo, Art Painting, Cartoon, and Sketch, including $7$ object categories and $9,991$ images total. We adopt the official split provided by \cite{li2017deeper} for training and validation.
(2) \textbf{Office-Home} \cite{venkateswara2017deep} contains around $15,500$ images of $65$ categories from $4$ domains: Artistic, Clipart, Product and Real-World. As in \cite{carlucci2019domain}, we randomly split each domain into $90\%$ for training and $10\%$ for validation.
(3) \textbf{VLCS} \cite{torralba2011unbiased} comprises of $5$ categories selected from $4$ domains, VOC $2007$ (Pascal), LabelMe, Caltech and Sun. We use the same setup as \cite{carlucci2019domain} and divide the dataset into training and validation sets based on $7:3$.
(4) \textbf{DomainNet} \cite{peng2019moment} is a large-scale dataset, consisting of about $586,575$ images with $345$ categories from $6$ domains, \ie,
Clipart, Infograph, Painting, Quickdraw, Real, and Sketch. Following \cite{gulrajani2020search}, we split source data into $80\%$ for training and $20\%$ for validation.

\subsection{Implementation Details} 

\textbf{Basic details.} 
For PACS and OfficeHome, we use the ImageNet pre-trained ResNet-$18$ and ResNet-$50$ as our backbones following \cite{zhang2022mvdg,meng2022attention}. 
For VLCS, we follow \cite{zhang2021deep,zhang2022mvdg} and use the ResNet-$18$ as backbone. 
The batch size is $128$.
We train the network using SGD with momentum of $0.9$ and weight decay of $0.0005$ for $50$ epochs. The initial learning rate is $0.002$ and decayed by $0.1$ at $80\%$ of the total epochs.
For the large-scale DomainNet, we use ResNet-$50$ pre-trained on ImageNet as backbone and train the network using Adam optimizer for $5000$ iterations following \cite{cha2022domain,gulrajani2020search}. 
The initial learning rate is $5e-5$ and the batch size is $64$. 

\textbf{Method-specific details.} 
For all experiments, we the dropout ratio $P_{drop}$ of DomainDrop to $0.33$. 
We set the weight of gradient reversal layers \cite{matsuura2020domain} before domain discriminators to $0.25$. 
The weight of the dual consistency loss 
is set to $1.5$ and the temperature $T$ in Eq.~(\ref{eq:consistency}) is set to $5$ for all datasets.
For layer-wise training scheme, we select all the middle layers of the network (\ie, the $1$st, $2$nd, $3$rd, and $4$-th residual blocks) as the candidate set. 
At each iteration, we randomly select a residual block from the set and perform DomainDrop on its feature maps. 
We apply the leave-one-domain-out protocol for all benchmarks.
We select the best model on the validation splits of all source domains and report the top-$1$ accuracy.
All results are reported based on the averaged accuracy over five repetitive runs.

\subsection{Comparison with SOTA Methods}
\textbf{Evaluation on PACS.} 
We evaluate our framework on PACS with ResNet-$18$ and ResNet-$50$ as our backbones.
We compare with several dropout-based methods (\ie, I$^{2}$-drop \cite{shi2020informative}, RSC \cite{huang2020self}, PLACE \cite{guo2021domain}, CDG \cite{du2022cross}), augmentation based methods (\ie, MixStyle \cite{zhou2021domain}, LDSDG \cite{wang2021learning}, EFDMix \cite{zhang2022exact}, StyleNeophile \cite{kang2022style}, ALOFT \cite{guo2023aloft}), feature decorrelation methods (\ie, StableNet \cite{zhang2021deep},  $\rm I^{2}$-ADR \cite{meng2022attention}), meta-learning method (\ie, COMEN \cite{chen2022compound}) and neural search (\ie, NAS-OoD \cite{bai2021ood}).
As presented in Tab.~\ref{table: PACS}, our framework obtains the highest average accuracy among all the compared methods on both backbones. 
Specifically, compared with existing dropout-based DG methods, our DomainDrop can surpass the SOTA approach CDG by a considerable margin by $0.86\%$ ($86.66\%$ vs. $85.80\%$) on ResNet-$18$ and $0.81\%$ ($89.51\%$ vs. $88.70\%$) on ResNet-$50$.
It is because DomainDrop can explicitly reduce the domain-specific features in every middle layer of the network, instead of only applying dropout in a single layer and narrowing the domain gap implicitly. 
Besides, our method achieves the best performance on most domains and our overall performance exceeds other SOTA DG methods. 
The encouraging results demonstrate the superiority of our method in fighting the overfitting issue of the model on source domains.

\begin{table}[tb!]
    \begin{center}
    \caption{Performance (\%) comparisons with the start-of-the-art DG approaches on the PACS dataset with ResNet-$18$ and ResNet-$50$ backbones.
    The best performance is marked as \textbf{bold}.
    }
    \label{table: PACS}
      \resizebox{\linewidth}{!}{
    \begin{tabular}{l | cccc |c}
        \hline
        Methods & Art & Cartoon & Photo & Sketch & Avg.\\
        \hline
        \multicolumn{6}{c}{ResNet-18} \\
        \hline
        DeepAll \cite{zhou2020deep} {\scriptsize (AAAI'20)} & 80.31 & 76.65 & 95.38 & 71.67 & 81.00 \\
        I$^{2}$-Drop \cite{shi2020informative} {\scriptsize (ICML'20)} & 80.27 & 76.54 & 96.11 & 76.38 & 82.33 \\
        DMG \cite{chattopadhyay2020learning} {\scriptsize (ECCV'20)} & 80.38 & 76.90 & 93.35 & 75.21 & 81.46 \\
        RSC \cite{huang2020self} {\scriptsize (ECCV'20)} & 83.43 & 80.31 & 95.99 & 80.85 & 85.15\\
        NAS-OoD \cite{bai2021ood} {\scriptsize (ICCV'21)} & 83.74 & 79.69 & 96.23 & 77.27 & 84.23 \\
        MixStyle \cite{zhou2021domain} {\scriptsize (ICLR'21)} & 84.10 & 78.80 & 96.10 & 75.90 & 83.70 \\
        LDSDG \cite{wang2021learning} {\scriptsize (ICCV'21)} & 81.44 & 79.56 & 95.51 & 80.58 & 84.27 \\
        PLACE \cite{guo2021domain} {\scriptsize (arXiv'21)} & 82.60 & 78.33 & 95.65 & 81.47 & 84.51 \\
        FACT \cite{xu2021fourier} {\scriptsize (CVPR'21)} & \textbf{85.37} & 78.38 & 95.15 & 79.15 & 84.51 \\
        StableNet \cite{zhang2021deep} {\scriptsize (CVPR'21)} & 81.74 & 79.91 & 96.53 & 80.50 & 84.69 \\
        EFDMix \cite{zhang2022exact} {\scriptsize (CVPR'22)} & 83.90 & 79.40 & 96.80 & 75.00 & 83.90 \\
        StyleNeophile \cite{kang2022style} {\scriptsize (CVPR'22)} & 84.41 & 79.25 & 94.93 & 83.27 & 85.47 \\
        COMEN \cite{chen2022compound} {\scriptsize (CVPR'22)} & 82.60 & \textbf{81.00} & 94.60 & 84.50 & 85.70 \\
        $\rm I^{2}$-ADR \cite{meng2022attention} {\scriptsize (ECCV'22)} & 82.90 & 80.80 & 95.00  & 83.50 & 85.60 \\
        CDG \cite{du2022cross} {\scriptsize (IJCV'22)} & 83.50 & 80.10 & 95.60 & 83.80 & 85.80 \\
        ALOFT \cite{guo2023aloft} {\scriptsize (CVPR'23)} & 84.81 & 79.05 & 96.11 & 80.55 & 85.13 \\ 
        \hline
        \rowcolor{mygray}
        DomainDrop {\scriptsize (Ours)} & 84.47{\tiny $\pm$ 0.77} & 80.50{\tiny $\pm$ 0.56} & \textbf{96.83}{\tiny $\pm$ 0.21} & \textbf{84.83}{\tiny $\pm$ 0.67} & \textbf{86.66} \\
        \hline
        \hline
        \multicolumn{6}{c}{ResNet-50} \\
        \hline
        DeepAll \cite{zhou2020deep} {\scriptsize (AAAI'20)} & 81.31 & 78.54 & 94.97 & 69.76 & 81.15 \\
        RSC \cite{huang2020self} {\scriptsize (ECCV'20)} & 87.89 & 82.16 & 97.92 & 83.85 & 87.83 \\
        FACT \cite{xu2021fourier} {\scriptsize (CVPR'21)} & 89.63 & 81.77 & 96.75 & 84.46 & 88.15 \\
        PLACE \cite{guo2021domain} {\scriptsize (arXiv'21)} & 87.55 & 83.11 & 97.19 & 83.48 & 87.83 \\
        COPA \cite{wu2021collaborative} {\scriptsize (ICCV'21)} & 83.30 & 79.80 & 94.60 & 82.50 & 85.10 \\
        SWAD \cite{cha2021swad} {\scriptsize (NeurIPS'21)} & 89.30 & 83.40 & 97.30 & 82.50 & 88.10 \\ 
        EFDMix \cite{zhang2022exact} {\scriptsize (CVPR'22)} & \textbf{90.60} & 82.50 & 98.10 & 76.40 & 86.90 \\
        StyleNeophile \cite{kang2022style} {\scriptsize (CVPR'22)} & 90.35 & 84.20 & 96.73 & 85.18 & 89.11 \\
        $\rm I^{2}$-ADR \cite{meng2022attention} {\scriptsize (ECCV'22)} & 88.50 & 83.20 & 95.20  & 85.80 & 88.20 \\
        PTE \cite{min2022grounding} {\scriptsize (ECCV'22)} & 87.90 & 78.40 & \textbf{98.20} & 75.70 & 85.10 \\
        CDG \cite{du2022cross} {\scriptsize (IJCV'22)} & 88.90 & 83.50 & 97.60 & 84.90 & 88.70 \\  
        ALOFT \cite{guo2023aloft} {\scriptsize (CVPR'23)} & 89.26 & 83.11 & 97.96 & 84.04 & 88.59 \\
        \hline
        \rowcolor{mygray} DomainDrop {\scriptsize (Ours)} & 89.82{\tiny $\pm$ 0.44} & \textbf{84.22}{\tiny $\pm$ 0.37} & 98.02{\tiny $\pm$ 0.24} & \textbf{85.98}{\tiny $\pm$ 1.14} & \textbf{89.51} \\
        \hline
    \end{tabular}}
    \end{center}
    \vspace{-0.45cm}
\end{table}

\begin{table}[tb!] 
    \begin{center}
    \caption{Performance (\%) comparisons with the SOTA DG methods on the OfficeHome dataset with ResNet-$18$ and ResNet-$50$ backbones.
    The best performance is marked as \textbf{bold}.
    }
    \label{table: OfficeHome}
      \resizebox{\linewidth}{!}{
    \begin{tabular}{l | cccc |c}
        \hline
        Methods & Art & Clipart & Product & Real & Avg.\\
        \hline
        \multicolumn{6}{c}{ResNet-18} \\
        \hline
        DeepAll \cite{zhou2020deep} {\scriptsize (AAAI'20)} & 56.98 & 50.14 & 72.85 & 74.39 & 63.59 \\
        RSC \cite{huang2020self} {\scriptsize (ECCV'20)} & 57.70 & 48.58 & 72.59 & 74.17 & 63.26 \\
        MixStyle \cite{zhou2021domain} {\scriptsize (ICLR'21)} & 58.70 & 53.40 & 74.20 & 75.90 & 65.50\\
        SagNet \cite{nam2021reducing} {\scriptsize (CVPR'21)} & 60.20 & 45.38 & 70.42 & 73.38 & 62.34 \\
        COPA \cite{wu2021collaborative} {\scriptsize (ICCV'21)} & 59.40 & 55.10 & 74.80 & 75.00 & 66.10 \\
        FACT \cite{xu2021fourier} {\scriptsize (CVPR'21)} & \textbf{60.34} & 54.85 & 74.48 & 76.55 & 66.56 \\
        StyleNeophile \cite{kang2022style} {\scriptsize (CVPR'22)} & 59.55 & 55.01 & 73.57 & 75.52 & 65.89 \\
        CDG \cite{du2022cross} {\scriptsize (IJCV'22)} & 59.20 & 54.30 & \textbf{74.90} & 75.70 & 66.00 \\
        \hline
        \rowcolor{mygray} DomainDrop {\scriptsize (Ours)} & 59.62{\tiny $\pm$ 0.40} & \textbf{55.60}{\tiny $\pm$ 0.31} & 74.50{\tiny $\pm$ 0.53} & \textbf{76.64}{\tiny $\pm$ 0.35} & \textbf{66.59} \\
        \hline
        \hline
        \multicolumn{6}{c}{ResNet-50} \\
        \hline
        DeepAll \cite{zhou2020deep} {\scriptsize (AAAI'20)} & 61.30 & 52.40 & 75.80 & 76.60 & 66.50 \\
        RSC \cite{huang2020self} {\scriptsize (ECCV'20)} & 57.70 & 51.40 & 74.80 & 75.10 & 65.50 \\
        SelfReg \cite{kim2021selfreg} {\scriptsize (ICCV'21)} & 63.60 & 53.10 & 76.90 & 78.10 & 67.90 \\
        SagNet \cite{nam2021reducing} {\scriptsize (CVPR'21)} & 63.40 & 54.80 & 75.80 & 78.30 & 68.10 \\
        SWAD \cite{cha2021swad} {\scriptsize (NeurIPS'21)} & 66.10 & 57.70 & 78.40 & 80.20 & 70.60 \\   
        Fishr \cite{rame2022fishr} {\scriptsize (ICML'22)} & 63.40 & 54.20 & 76.40 & 78.50 & 68.20 \\
        PTE \cite{min2022grounding} {\scriptsize (ECCV'22)} & 66.30 & 55.80 & 78.20 & \textbf{80.40} & 70.10 \\
        \hline
        \rowcolor{mygray} DomainDrop {\scriptsize (Ours)} & \textbf{67.33}{\tiny $\pm$ 0.45} & \textbf{60.39}{\tiny $\pm$ 0.48} & \textbf{79.05}{\tiny $\pm$ 0.29} & 80.22{\tiny $\pm$ 0.22} & \textbf{71.75}\\
        \hline
    \end{tabular}}
    \end{center}
    \vspace{-0.35cm}
\end{table}


\textbf{Evaluation on OfficeHome.} 
We also compare our method with SOTA DG methods on OfficeHome to demonstrate the adaptation of our method to the dataset with a large number of categories and samples. 
We perform the evaluation on both ResNet-$18$ and ResNet-$50$ backbones.
The results are reported in Tab.~\ref{table: OfficeHome}. 
Our method can still achieve the best performance among the compared DG methods, \eg, outperforming the state-of-the-are DG method COPA \cite{wu2021collaborative} by $0.49\%$ ($66.59\%$ vs $66.10\%$).
Besides, our method precedes the best method SWAD \cite{cha2021swad} on ResNet-$50$, which proposes an ensemble learning method that seeks flat minima for DG, with a significant improvement of $1.14\%$ ($71.75\%$ vs. $70.60\%$).
The above results further demonstrate the effectiveness of our framework.

\textbf{Evaluation on VLCS.} 
To verify the trained model can also generalize to unseen target domains with a relatively small domain gap to source domains, we conduct the experiments on VLCS with ResNet-$18$.
The results are portrayed in Tab.~\ref{table: VLCS}. 
Among all the competitors, our DomainDrop achieves the best performance, exceeding the second-best method StableNet \cite{zhang2021deep} by $0.60\%$ ($78.25\%$ vs. $77.65\%$) on average.
Our method also exceeds the advanced dropout method RSC \cite{huang2020self}, which removes over-dominate features according to gradients, by $2.36 \%$ ($78.25\%$ vs. $75.89\%$).
The results prove the effectiveness of our method again.

\begin{table}[tb!]
    \begin{center}
    \caption{Performance (\%) comparisons with the SOTA DG methods on VLCS with ResNet-$18$ backbone.
    The best is \textbf{bolded}.}
    \label{table: VLCS}
    \resizebox{\linewidth}{!}{
    \begin{tabular}{l | cccc |c}
    \hline
    Methods & Caltech & LabelMe & Pascal & Sun & Avg.\\
    \hline
    DeepAll \cite{zhou2020deep} {\scriptsize (AAAI'20)} & 96.98 & 62.00 & 73.83 & 68.66 & 75.37 \\
    JiGen \cite{carlucci2019domain} {\scriptsize (CVPR'19)} & 96.17 & 62.06 & 70.93 & 71.40 & 75.14 \\
    MMLD \cite{matsuura2020domain} {\scriptsize (AAAI'20)} & 97.01 & 62.20 & 73.01 & 72.49 & 76.18\\
    RSC \cite{huang2020self} {\scriptsize (ECCV'20)} & 95.83 & 63.74 & 71.86 & 72.12 & 75.89 \\
    StableNet \cite{zhang2021deep} {\scriptsize (CVPR'21)} & 96.67 & \textbf{65.36} & 73.59 & \textbf{74.97} & 77.65\\
    \hline
    \rowcolor{mygray} DomainDrop {\scriptsize (Ours)} & \textbf{98.94}{\tiny $\pm$ 0.19} & 63.97{\tiny $\pm$ 1.33} & \textbf{76.36}{\tiny $\pm$ 0.93} & 73.74{\tiny $\pm$ 1.17} & \textbf{78.25} \\
    \hline
\end{tabular}}
\end{center}
\vspace{-0.5cm}
\end{table}


\textbf{Evaluation on DomainNet.} 
Tab.~\ref{table: DomainNet} shows the results on the large-scale DomainNet. 
On the challenging benchmark, our method performs better in the averaged accuracy than existing methods and improves top-$1$ accuracy by $3.87\%$ ($44.37\%$ vs. $40.50\%$) from ResNet-$50$ baseline, proving the effectiveness of our method on the large-scale dataset.
All the comparisons prove that suppressing domain-sensitive channels can effectively improve model generalizability.

\begin{table}[tb!]
    \begin{center}
    \caption{Performance (\%) comparisons with the SOTA DG methods on DomainNet with ResNet-$50$ backbone.
    The best is \textbf{bolded}.}
    \label{table: DomainNet}
    \resizebox{\linewidth}{!}{
    \begin{tabular}{l | cccccc |c}
    \hline
    Methods & Clipart & Infograph & Painting & Quickdraw & Real & Sketch  & Avg.\\
    \hline
    DeepAll \cite{zhou2020deep} {\scriptsize (AAAI'20)} & 62.20 & 19.90 & 45.50 & 13.80 & 57.50 & 44.40 & 40.50 \\
    RSC \cite{huang2020self} {\scriptsize (ECCV'20)} & 55.00 & 18.30 & 44.40 & 12.20 & 55.70 & 47.80 & 38.90 \\
    SagNet \cite{nam2021reducing} {\scriptsize (CVPR'21)} & 57.70 & 19.00 & 45.30 & 12.70 & 58.10 & 48.80 & 40.30 \\
    SelfReg \cite{kim2021selfreg} {\scriptsize (ICCV'21)} & 60.70 & 21.60 & 49.40 & 12.70 & 60.70 & 51.70 & 42.80 \\
    I$^{2}$-ADR \cite{meng2022attention} {\scriptsize (ECCV'22)} & \textbf{64.40} & 20.20 & 49.20 & \textbf{15.00} & 61.60 & 53.30 & 44.00 \\
    PTE \cite{min2022grounding} {\scriptsize (ECCV'22)} & 62.40 & 21.00  & 50.50 & 13.80 & \textbf{64.60} & 52.40 & 44.10 \\
    \hline
    \rowcolor{mygray} DomainDrop {\scriptsize (Ours)} & 62.93{\tiny $\pm$ 0.25} & \textbf{21.63}{\tiny $\pm$ 0.07} & \textbf{50.67}{\tiny $\pm$ 0.15} & 14.83{\tiny $\pm$ 0.30} & 62.70{\tiny $\pm$ 0.08} & \textbf{53.46}{\tiny $\pm$ 0.59} & \textbf{44.37}\\
    \hline
\end{tabular}}
\end{center}
\vspace{-0.2cm}
\end{table}


\begin{table}[tb!]
    \begin{center}
    \caption{
        Comparisons with the SOTA dropout-based methods on PACS dataset with ResNet-$18$ backbone. The best is \textbf{bolded}.
    }
    \label{table: dropout}
      \resizebox{\linewidth}{!}{
    \begin{tabular}{l | cccc |c}
        \hline
        Methods & Art & Cartoon & Photo & Sketch & Avg.\\
        \hline
        Baseline {\scriptsize (AAAI'20)} & 80.31 & 76.65 & 95.38 & 71.67 & 81.00 \\
        Cutout \cite{devries2017improved} {\scriptsize (ArViv'17)} & 79.60 & 77.20 & 95.90 & 71.60 & 80.60 \\
        C-Drop \cite{morerio2017curriculum} {\scriptsize (ICCV'17)} & 79.64 & 76.49 & 95.93 & 72.37 & 81.11 \\
        DropBlock \cite{ghiasi2018dropblock} {\scriptsize (NeurIPS'18)} & 80.30 & 77.50 & 95.60 & 76.40 & 82.50 \\
        AdvDrop \cite{park2018adversarial} {\scriptsize (AAAI'18)} & 82.40 & 78.20 & 96.10 & 75.90 & 83.00 \\
        WCD \cite{hou2019weighted} {\scriptsize (AAAI'19)} & 81.56 & 78.24 & 94.99 & 75.53 & 82.58 \\
        DMG \cite{chattopadhyay2020learning} {\scriptsize (ECCV'20)} & 80.38 & 76.90 & 93.35 & 75.21 & 81.46 \\
        I$^{2}$-Drop \cite{shi2020informative} {\scriptsize (ICML'20)} & 80.27 & 76.54 & 96.11 & 76.38 & 82.33 \\
        RSC \cite{huang2020self} {\scriptsize (ECCV'20)} & 83.43 & 80.31 & 95.99 & 80.85 & 85.15\\
        PLACE \cite{guo2021domain} {\scriptsize (arXiv'21)} & 82.60 & 78.33 & 95.65 & 81.47 & 84.51 \\
        CDG \cite{du2022cross} {\scriptsize (IJCV'22)} & 83.50 & 80.10 & 95.60 & 83.80 & 85.80 \\
        \hline
        \rowcolor{mygray}
        DomainDrop {\scriptsize (Ours)} &  \textbf{84.47}{\tiny $\pm$ 0.77} & \textbf{80.50}{\tiny $\pm$ 0.56} & \textbf{96.83}{\tiny $\pm$ 0.21} & \textbf{84.83}{\tiny $\pm$ 0.67} & \textbf{86.66} \\
        \hline
    \end{tabular}}
    \end{center}
    \vspace{-0.5cm}
\end{table}

\textbf{Comparison with SOTA dropout methods.} 
We here compare DomainDrop with the SOTA dropout-based methods, including the traditional approaches designed for supervised learning (\ie, Cutout \cite{devries2017improved}, C-Drop \cite{morerio2017curriculum}, DropBlock \cite{ghiasi2018dropblock}, AdvDrop \cite{park2018adversarial}, WCD \cite{hou2019weighted}) and the methods proposed for domain generalization (\ie, DMG \cite{chattopadhyay2020learning}, I$^{2}$-Drop \cite{shi2020informative}, RSC \cite{huang2020self}, PLACE \cite{guo2021domain}, CDG \cite{du2022cross}). 
The results are summarized in Tab.~\ref{table: dropout}.
We observe that the traditional dropout methods cannot adequately combat the overfitting issue of the model on source domains, which is caused by the large discrepancy between source domains and unseen target domains.
Besides, DMG \cite{chattopadhyay2020learning} that assumes some domain-specific features provide useful information for the target domain achieves a slight improvement, which may be because the assumption cannot always be guaranteed since the target domain is unseen.
I$^{2}$-Drop \cite{shi2020informative} aims to penalize the model from learning superficial local textures, but there could also exist other forms of domain-specific features. 
PLACE \cite{guo2021domain} attempts to activate diverse channels by randomly discarding channels, but it inevitably activates channels that capture domain-specific information, impeding the model generalization.
Although RSC \cite{huang2020self} and CDG \cite{du2022cross} have shown promising performance, these methods are only suitable for a single network layer, which prevents them from sufficiently reducing domain-specific information.
In general, our framework obtains the best performance among existing dropout methods, which proves its effectiveness in enhancing model generalization.

\vspace{0.05cm}

\subsection{Analytical Experiments}
In this paragraph, we conduct extensive ablation studies of our framework, including the impact of each component and the parameter sensitivity. 
We also analyze the discrepancy among source and target domains to prove that our framework can effectively reduce domain divergence.
Moreover, we validate the orthogonality of our DomainDrop on other DG SOTA methods.
More experiments can be found in the Supplementary Material.


\textbf{Ablation study on each component.} 
We here conduct the ablation study to investigate the efficacy of DomainDrop (DD), Layer-wise Training Scheme (LT), and Consistency Loss (CL) in our framework.
Tab.~\ref{tab:components} presents the results of different variants of DomainDrop with ResNet-$18$ on both PACS and OfficeHome. 
Notably, for \textit{Variant $1$}, we apply DomainDrop to all layers, activating it with a probability of $0.5$ for each layer, which aligns with \cite{zhou2021domain,zhang2022exact}. 
As shown in Tab.~\ref{tab:components}, variant $1$ exhibits significant improvement over the baseline, \ie, $4.06\%$ ($85.06\%$ vs. $81.00\%$) on PACS and $1.71\%$ ($65.30\%$ vs. $63.59\%$) on OfficeHome, proving the effectiveness of DomainDrop in reducing channel sensitivity to domain changes and narrowing domain gap.
Besides, comparing variant $1$ with variant $2$, we observe that combining both DomainDrop and layer-wise training scheme leads to better performance than using DomainDrop alone, indicating that alternating use of DomainDrop at multiple layers can maximize regularization effect while avoiding losing too much information. 
Furthermore, the improved performance of variant $3$ over variant $1$ suggests that the consistency loss contributes to enhancing domain-invariant features.
Finally, DomainDrop performs the best, verifying that the three modules complement and promote mutually, and none of them is dispensable for achieving the superior generalization ability of the model.

\begin{table}[tb!] 
    \begin{center}
    \caption{
        Ablation study on different components of our framework: DomainDrop (DD), layer-wise training scheme (LT), and dual consistency loss (CL).
        The experiment is conducted on PACS and OfficeHome with ResNet-$18$.
        The best is \textbf{bolded}.
    }
    \vspace{0.4cm}
    \label{table: PACS}
      \resizebox{\linewidth}{!}{
    \begin{tabular}{l | ccc | cccc |c}
        \hline
        \multicolumn{9}{c}{PACS} \\
        \hline
        Methods & DD & LT & CL & Art & Cartoon & Photo & Sketch & Avg.\\
        \hline
        Baseline & - & - & - & 80.31{\tiny $\pm$ 1.54} & 76.65{\tiny $\pm$ 0.48} & 95.38{\tiny $\pm$ 0.12} & 71.67{\tiny $\pm$ 1.49} & 81.00 \\
        \hline
        Variant 1 & \checkmark & - & - & 82.37{\tiny $\pm$ 0.78} & 79.27{\tiny $\pm$ 0.26} & 95.45{\tiny $\pm$ 0.27} & 83.15{\tiny $\pm$ 0.73} & 85.06 \\
        Variant 2 & \checkmark & \checkmark & - & 83.84{\tiny $\pm$ 0.12} & 79.69{\tiny $\pm$ 0.29} & 96.47{\tiny $\pm$ 0.36} & 83.74{\tiny $\pm$ 0.39} & 85.94 \\
        Variant 3 & \checkmark & - & \checkmark & 83.32{\tiny $\pm$ 0.76} & 80.09{\tiny $\pm$ 0.91} & 95.85{\tiny $\pm$ 0.33} & 83.71{\tiny $\pm$ 0.61} & 85.74 \\
        \hline
        \rowcolor{mygray} DomainDrop & \checkmark & \checkmark & \checkmark &  \textbf{84.47}{\tiny $\pm$ 0.77} & \textbf{80.50}{\tiny $\pm$ 0.56} & \textbf{96.83}{\tiny $\pm$ 0.21} & \textbf{84.83}{\tiny $\pm$ 0.67} & \textbf{86.66} \\
        \hline
        \hline
        \multicolumn{9}{c}{OfficeHome} \\
        \hline
        Methods & DD & LT & CL & Art & Clipart & Product & Real & Avg.\\
        \hline
        Baseline & - & - & - & 56.98{\tiny $\pm$ 0.47} & 50.14{\tiny $\pm$ 0.88} & 72.85{\tiny $\pm$ 0.35} & 74.39{\tiny $\pm$ 0.32} & 63.59 \\
        \hline
        Variant 1 & \checkmark & - & - & 57.49{\tiny $\pm$ 0.37} & 53.69{\tiny $\pm$ 0.58} & 73.82{\tiny $\pm$ 0.14} & 76.18{\tiny $\pm$ 0.20} & 65.30\\
        Variant 2 & \checkmark & \checkmark & - & 59.46{\tiny $\pm$ 0.28} & 53.86{\tiny $\pm$ 0.57} & 74.23{\tiny $\pm$ 0.29} & 76.31{\tiny $\pm$ 0.24} & 65.96 \\
        Variant 3 & \checkmark & - & \checkmark & 58.17{\tiny $\pm$ 0.29} & 54.96{\tiny $\pm$ 0.40} & 73.95{\tiny $\pm$ 0.35} & 76.42{\tiny $\pm$ 0.49} & 65.88 \\
        \hline
        \rowcolor{mygray} DomainDrop & \checkmark & \checkmark & \checkmark & \textbf{59.62}{\tiny $\pm$ 0.40} & \textbf{55.60}{\tiny $\pm$ 0.31} & \textbf{74.50}{\tiny $\pm$ 0.53} & \textbf{76.64}{\tiny $\pm$ 0.35} & \textbf{66.59}\\
        \hline
    \end{tabular}}
    \label{tab:components}
    \end{center}
    \vspace{-0.6cm}
\end{table}

\begin{figure}[tb!]
    \centering
    \hspace{-0.6cm}
    \subfigure[Effect of dropout probability.]{
        \begin{minipage}[t]{0.45\linewidth}
        \centering
        \includegraphics[scale=0.27]{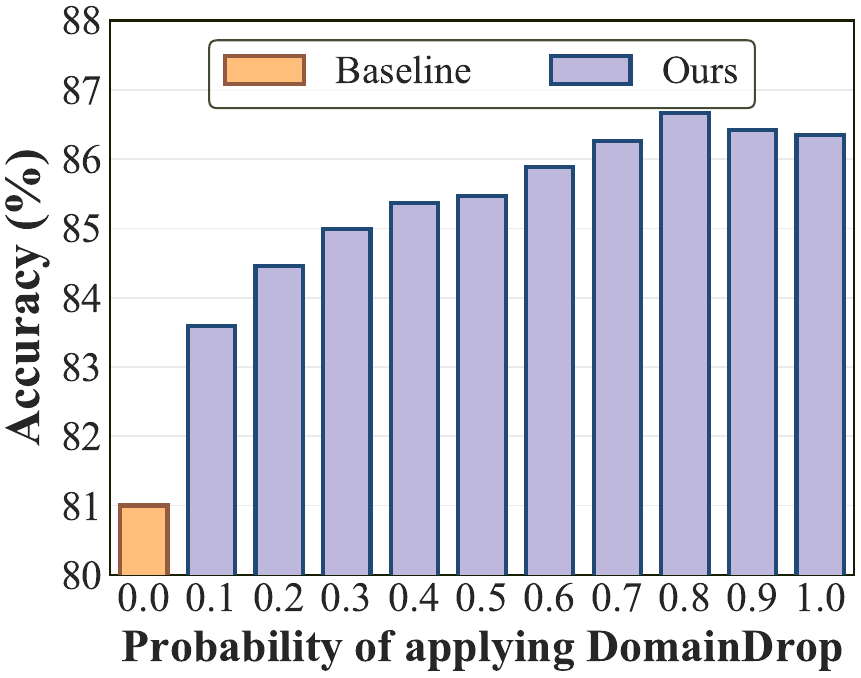}
        \label{fig:probability}
        \vspace{-0.4cm}
    \end{minipage}}
        \hspace{0.2cm}
    \subfigure[Effects of dropout ratio.]{
        \begin{minipage}[t]{0.45\linewidth}
        \centering
        \includegraphics[scale=0.27]{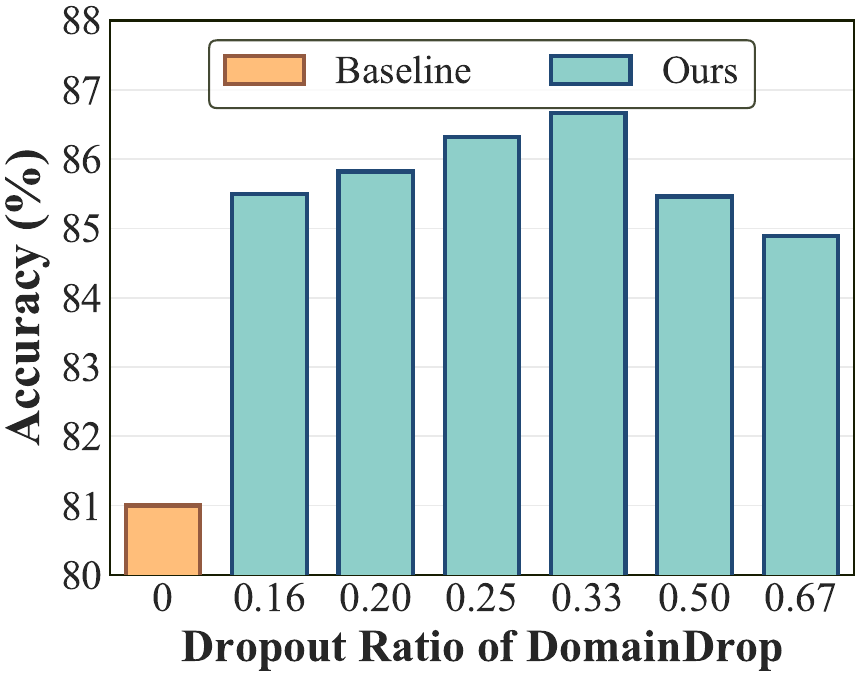}
        \label{fig:ratio}
        \vspace{-0.4cm}
    \end{minipage}}
      \vspace{0.05cm}
    \caption{
        Effects of hyper-parameters including the probability of applying DomainDrop
        and the dropout ratio in DomainDrop. The experiments are conducted
        on PACS with ResNet-$18$ backbone.}
    \label{fig:hyper}
\end{figure}

\textbf{Parameter sensitivity.} 
We conduct experiments to investigate the sensitivity of DomainDrop to hyper-parameter $P_{active}$ and $P_{drop}$ as shown in Fig.~\ref{fig:probability} and \ref{fig:ratio}.
Specifically, $P_{apply}$ denotes the probability of applying DomainDrop to the network at each iteration, varying from $0.1$ to $1.0$. 
The results in Fig.~\ref{fig:probability} show that DomainDrop achieves competitive performance robustly with a relatively large probability (\ie, $P_{active} \geq 0.7$), verifying the stability of our method.
Moreover, we also examine the impacts of the dropout ratio on model performance, which vary $\{0.16, 0.20, 0.25, 0.33, 0.50, 0.67\}$ as in \cite{huang2020self}.
Fig.~\ref{fig:ratio} shows that our framework consistently exceeds the baseline by a large margin (\ie, more than $4\%$) with different dropout ratios, and the highest accuracy is reached at $P_{drop} = 0.33$, which is adopted as default in all our experiments.

\begin{table}[tb!]
    \centering
    \caption{
        Comparisons of source and source-target domain divergences ($\times 10$) for different methods on PACS with ResNet-$18$.
    }
    \vspace{0.25cm}
    \resizebox{\linewidth}{!}{
        \setlength{\tabcolsep}{3mm}{
    \begin{tabular}{l|cc}
    \hline
    Methods & Source Divergence & Source-Target Divergence \\
    \hline
    Baseline & 9.38 & 7.25 \\
    MixStyle \cite{zhou2021domain} & 7.53 & 6.23 \\
    FACT \cite{xu2021fourier} & 9.00 & 7.27 \\
    I$^2$-Drop \cite{meng2022attention} & 7.11 & 6.44 \\
    RSC \cite{huang2020self} & 8.81 & 7.20 \\
    PLACE \cite{guo2021domain} & 6.45 & 6.17 \\
    \hline
    \rowcolor{mygray} DomainDrop (Ours) & \textbf{5.15} & \textbf{4.09} \\
    \hline
    \end{tabular}
    }
    }
    \label{tab:estimated CMMD}
    \vspace{-0.2cm}
\end{table}

\textbf{Discrepancy among source and target domains.} 
\label{sec:prove theory}
To verify the effectiveness of our method to channel deviation across domains, we here measure the estimated channel-level maximum mean discrepancy (CMMD) among source and target domains. 
Based on the definition of CMMD in Eq.~(\ref{eq:CMMD paper}), given a trained feature extractor $\phi: \mathcal{X} \rightarrow \mathbb{R}^n$, we estimate the CMMD between pairwise domains $D_s^i$ and $D_s^j$: 
\vspace{-0.3cm}
\begin{equation}
    \begin{aligned}
        \hat{d}_{CMMD}(D_s^i, D_s^j)=\frac{1}{n} \sum_{k=1}^{n}||\phi_k(D_s^i) - \phi_k(D_s^j)||_2.
        \label{eq:CMMD estimated}
    \end{aligned}
\end{equation}
Then we estimate the source domain divergence $\beta$ by $\hat{\beta} = sup_{i, j \in [K]} \hat{d}_{\text{CMMD}}(D_s^i, D_s^j)$. 
We utilize the averaged divergence between target domain and each source domain to estimate the source-target domain gap $\gamma$, defined as $\hat{\gamma} = \frac{1}{K} \sum_{k=1}^{K} \hat{d}_{\text{CMMD}}(D_s^k, D_t)$. 
The results are shown in Tab.~\ref{tab:estimated CMMD}. 
Notably, both the SOTA DG methods, MixStyle \cite{zhou2021domain} and FACT \cite{xu2021fourier}, present lower source domain divergence due to their ability to diversify the source data.
However, they do not explicitly remove domain-specific features, leading to limited reduction of the source-target domain divergence. 
RSC \cite{huang2020self} regularizes the model to learn comprehensive representations but still cannot adequately narrow the domain gap across source and target domains.
I$^{2}$-Drop \cite{shi2020informative} penalizes superficial local features that may contain domain-specific information, but it only applies to shallow layers and cannot suppress other forms of domain-specific features.
PLACE \cite{guo2021domain} generates a regularization effect to fight the overfitting, but it cannot prevent channels from learning domain-specific information.
In contrast, the lowest source and source-target domain gaps achieved by DomainDrop prove its effectiveness in reducing domain-specific features, highlighting its superiority over existing methods. 
Moreover, the results align well with the theoretical analysis in Sec.~\ref{sec:theory} and prove that DomainDrop effectively lowers the generalization risk bound by reducing $\beta$ and $\gamma$.

\textbf{Orthogonality to other DG methods.} 
We investigate the effectiveness of DomainDrop in improving the performance of other DG methods. 
We selected two representative DG methods, \ie, RandAug \cite{cubuk2020randaugment} and FACT \cite{xu2021fourier}, both using ResNet-$18$ as the backbone. 
We also verify the effectiveness of DomainDrop on the SOTA MLP-like model GFNet \cite{rao2021global}.
As shown in Tab.~\ref{table:SOTA}, our method can significantly improve the SOTA DG methods, \eg, boosting RandAug by $3.06\%$ ($87.24\%$ vs. $84.18\%$) and FACT by $2.63\%$ ($87.24\%$ vs. $84.61\%$). 
These experiments demonstrate that our DomainDrop is orthogonal to other SOTA DG methods, and a new SOTA performance can be achieved by combining our approach with existing methods.
Moreover, our DomainDrop also achieves a considerable improvement on the MLP-like model, enhancing the performance of GFNet by $3.26\%$ ($91.02\%$ vs. $87.76\%$), which indicates the generalization of our methods on different network architectures.

\begin{table}[tb!]
    \begin{center}
    \caption{
        Effect (\%) of DomainDrop on other SOTA DG methods. 
        The experiments are conducted on the PACS dataset. 
        We select ResNet-$18$ as the backbone architecture for RandAug \cite{cubuk2020randaugment} and FACT \cite{xu2021fourier}.
        We also validate the effectiveness of our DomainDrop on the SOTA MLP-like model, \ie, GFNet \cite{rao2021global}.
    }
    \vspace{0.3cm}
    \label{table:SOTA}
      \resizebox{\linewidth}{!}{
        \setlength{\tabcolsep}{2.5mm}{
    \begin{tabular}{l | cccc |c}
        \hline
        Methods & Art & Cartoon & Photo & Sketch & Avg.\\
        \hline
        RandAug \cite{cubuk2020randaugment} & 82.90{\tiny $\pm$ 0.49} & 76.89{\tiny $\pm$ 0.75} & 96.17{\tiny $\pm$ 0.46} & 78.55{\tiny $\pm$ 0.54} & 83.63 \\
        \rowcolor{mygray} \textbf{+ DomainDrop} & \textbf{84.62}{\tiny $\pm$ 0.59}  & \textbf{80.25}{\tiny $\pm$ 0.51}  & \textbf{96.27}{\tiny $\pm$ 0.40}  & \textbf{85.62}{\tiny $\pm$ 0.60}  & \textbf{86.69} \\
        \hline
        FACT \cite{xu2021fourier} & 85.64{\tiny $\pm$ 0.57} & 77.94{\tiny $\pm$ 0.83} & 95.45{\tiny $\pm$ 0.78} & 79.41{\tiny $\pm$ 1.30} & 84.61 \\
        \rowcolor{mygray} \textbf{+ DomainDrop} & \textbf{86.52}{\tiny $\pm$ 0.83} & \textbf{80.86}{\tiny $\pm$ 0.49} & \textbf{95.81}{\tiny $\pm$ 0.23} & \textbf{85.75}{\tiny $\pm$ 1.02} & \textbf{87.24}\\
        \hline
        GFNet \cite{rao2021global} & 89.37{\tiny $\pm$ 0.60} & 84.74{\tiny $\pm$ 0.59} & 97.94{\tiny $\pm$ 0.25} & 79.01{\tiny $\pm$ 0.77} & 87.76
        \\
        \rowcolor{mygray} \textbf{+ DomainDrop} & \textbf{92.38}{\tiny $\pm$ 0.69} & \textbf{87.24}{\tiny $\pm$ 0.71} & \textbf{98.26}{\tiny $\pm$ 0.36} & \textbf{86.18}{\tiny $\pm$ 1.02} & \textbf{91.02} \\
        \hline
    \end{tabular}}
      }
      \vspace{-0.6cm}
    \end{center}
\end{table}

\vspace{-0.1cm}
\section{Conclusion}
In this paper, we study the model generalizability from a novel channel-level perspective and find that the overfitting issue could be caused by substantial domain-sensitive channels in the model. 
To tackle this issue, we propose a novel DG framework that explicitly suppresses domain-specific features by removing the domain-sensitive channels. 
We also theoretically prove the effectiveness of our framework to generate a tight generalization error bound. 
Experiments show that our framework achieves strong performance on various datasets compared with existing DG methods.


\clearpage

{\small
\bibliographystyle{ieee_fullname}
\bibliography{egbib}
}
\clearpage

\renewcommand\thesection{\Alph{section}}
\setcounter{section}{0}

\section{Theoretical Proof of Theorem 1}
\subsection{Notations} 
Given $K$ source domains $\mathcal{D}_s = \{D_s^1, D_s^2, ..., D_s^K\}$, we indicate that each domain $D_s^k$ contains $n_k$ input and labels $\{(x_i^k, y_i^k)\}_{i=1}^{n_k}$, where $x \in \mathcal{X}$ and $y \in \mathcal{Y}$.
The target domain is denoted as $\mathcal{D}_t$. 
Given a hypothesis $h: \mathcal{X} \rightarrow \mathcal{Y}$, where $h$ is from the space of the candidate hypothesis $\mathcal{H}$.
The expected risk of $h$ on a domain $D$ is defined as: $\mathcal{R}[h] = \mathbb{E}_{x \sim D} \ell[h(x), f(x)]$, where $\ell_{h, f}: x\rightarrow \ell[h(x), f(x)]$ is a convex loss-function defined for $\forall h, f \in \mathcal{H}$ and assumed to obey the triangle inequality.
Under the DG set, $y = f(x)$ represents the input label. 
We also denote the feature extractor of the network as $\phi(\cdot): \mathcal{X}\rightarrow \mathbb{R}^{n}$, which maps the input images into the $n$-dimentional feature space.
Following \cite{ding2022domain,zhao2022test}, for the source domains $\mathcal{D}_s = \{D_s^1, D_s^2, ..., D_s^K\}$, we define the convex hull $\Lambda_s$ as a set of mixture of source domain distributions: 
$\Lambda_s=\{\bar{D}: \bar{D}(\cdot) = \sum_{i=1}^K \pi_i D_s^i(\cdot), \pi_i \in \Delta_K\}$, where $\pi$ is non-negative coefficient in the $K$-dimensional simplex $\Delta_K$. 
We define $\bar{D}_t \in \Lambda_s$ as the closest domain to the target domain $D_t$.

\subsection{Definitions and Lemmas}

\noindent \textbf{Definition~1 \cite{redko2020survey}.} \textit{ 
    Let $\mathcal{F} = \{f\in \mathcal{H}_k: ||f||_{\mathcal{H}_k} \leq 1 \}$ be a function class, where $\mathcal{H}_k$ be a RKHS with its associated kernel $k$. Given two different distributions of $D_s$ and $D_t$, the maximum mean discrepancy (MMD) distance is:} 
\begin{equation}
    \begin{aligned}
        d_{\text{MMD}}(D_s, D_t)= ||\int_x k(x, \cdot)d(\phi(D_s)
        - \phi(D_t))||_{\mathcal{H}_k}.
        \label{eq:MMD}
    \end{aligned}
\end{equation}

Based on the MMD distance, we now introduce learning bounds for the target error where the divergence between distributions is measured by the MMD distance. 
We first introduce a lemma that indicates how the target error can be bounded by the empirical estimate of the MMD distance between an arbitrary pair of source and target domains.

\noindent \textbf{Lemma~1 \cite{redko2020survey}.} 
\label{Lemma 1} 
\textit{
Let $\mathcal{F} = \{f\in \mathcal{H}_k: ||f||_{\mathcal{H}_k} \leq 1 \}$ denote a function class, where $\mathcal{H}_k$ be a RKHS with its associated kernel $k$. 
Let $\ell_{h, f}: x\rightarrow \ell[h(x), f(x)]$ be a convex loss-function with a parameter form $|h(x) - f(x)|^q$ for some $q > 0$, and defined $\forall h, f \in \mathcal{F}$, $\ell$ obeys the triangle inequality. Let $S$ and $T$ be two samples of size $m$ drawn i.i.d from $D_s$ and $D_t$, respectively. Then, with probability of at least $1-\delta$ ($\delta \in (0, 1)$) for all $h \in \mathcal{F}$, the following holds:
} 
\begin{equation}
    \begin{aligned}
        \mathcal{R}_t[h] \leq & \mathcal{R}_s[h] + d_{\text{MMD}}(D_t, D_s) + \frac{2}{m} (
        E_{x \sim D_s}[\sqrt{tr(K_{D_s})}] \\
        & + E_{x\sim D_t}[\sqrt{tr(K_{D_t})}]) + 
        2\frac{\log(\frac{2}{\sigma})}{2m} + \epsilon,
    \end{aligned}
\end{equation}
\noindent \textit{
where $K_{D_s}$ and $K_{D_t}$ are kernel functions computed on samples from $D_s$ and $D_t$, respectively. $\epsilon$ is the combined error of the ideal hypothesis $h^*$ on $D_s$ and $D_t$.}

Then, to investigate the effect of channel robustness to domain shifts on the generalization error bound, we define the channel-level maximum mean discrepancy (CMMD) distance to estimate the channel-level distribution gap between different domains, which is formulated as:

\noindent \textbf{Definition~2.} \textit{Let $n$ denote the number of channels in the extracted features of $\phi(\cdot)$. Given two different distribution of $D_s$ and $D_t$, the channel-level maximum mean discrepancy (CMMD) between $\phi(D_s)$ and $\phi(D_t)$ is defined as:} 
\begin{equation}
    \begin{aligned}
        d_{\text{CMMD}}(D_s, D_t)=\frac{1}{n} \sum_{i=1}^{n}\sup_{\phi_i \in \Phi_i}||\int_x &k(x, \cdot)d(\phi_i(D_s) \\
        &- \phi_i(D_t))||_{\mathcal{H}_k},
        \label{eq:CMMD}
    \end{aligned}
\end{equation}
\noindent \textit{
where $\Phi$ is the space of candidate hypothesis for each channel, $\phi_i(D)$ is the distribution of the $i$-th channel for the domain $D$, and $\mathcal{H}_k$ is a RKHS with its associated kernel $k$.} 

The CMMD distance could be regarded as a channel-level version of the MMD distance, which represents the maximum value of the difference in channel activation for a given two domains in the model, thus reflecting the channel robustness to domain shifts.
Based on the CMMD distance and Lemma $1$, we derive a generalization error boundary of the model in the multi-source domain scenario (\ie, Theorem~$1$), and provide the detailed proof below.

\subsection{Proof}
\noindent \textbf{Theorem~1} \textbf{(Generalization risk bound)}. \textit{
    With the previous settings and assumptions, let $S^i$ and $T$ be two samples of size $m$ drawn i.i.d from $D_s^i$ and $D_t$, respectively.
    Then, with the probability of at least $1-\delta$ ($\delta \in (0, 1)$) for all $h \in \mathcal{F}$, the following inequality holds for the risk $\mathcal{R}_t[h]$:
}
\begin{equation}
    \begin{aligned}
        \mathcal{R}_t[h] \leq & \sum_{i=1}^{N}\pi_i\mathcal{R}_s^i[h] + d_{\text{CMMD}}(\bar{D}_t, D_t) \\
        & + \sup_{i, j \in [K]} d_{\text{CMMD}}(D_s^i, D_s^j) 
        + \lambda + \epsilon,
    \label{eq:generalization bound}
    \end{aligned}
\end{equation}
\textit{
    where $\lambda = 2\sqrt{\frac{\log(\frac{2}{\sigma})}{2m}} + \frac{2}{m} (\sum_{i=1}^{N}\pi_i 
    E_{x \sim D_s^i}[\sqrt{tr(K_{D_s^i})}] + E_{x\sim D_t}[\sqrt{tr(K_{D_t})}])$, $K_{D_s^i}$ and $K_{D_t}$ are kernel functions computed on samples from $D_s^i$ and $D_t$, respectively. $\epsilon$ is the combined error of ideal hypothesis $h^*$ on $D_t$ and $\bar{D}_t$.
}

\noindent\textbf{Proof.} Consider the closest domain $\bar{D}_t$ to target domain $D_t$ as a mixture distribution of $K$ source domains where the mixture weight is given by $\pi$, \ie, $\bar{D}_t = \sum_{i=1}^K \pi_i D_s^i(\cdot)$ with $\sum_{i=1}^K \pi_i = 1$. 
For a pair of source domain $D_s^i$ and the target domain $D_t$, the following inequality holds:
\begin{equation}
    \begin{aligned}
        d_{\text{CMMD}}(D_t, D_s^i) \leq d_{\text{CMMD}}(D_t, \bar{D_t}) + d_{\text{CMMD}}(\bar{D_t}, D_s^i).
    \end{aligned}
\end{equation}
According to Definition~$2$, we could derive the weighted sum of the CMMD distance between source domains and the target domain, which is formulated as:
\begin{equation}
    \begin{aligned}
        \sum_{i=1}^{N}\pi_i &d_{\text{CMMD}}(D_t, D_s^i) \\ 
        & \leq d_{\text{CMMD}}(D_t, \bar{D_t}) + \sum_{i=1}^{N}\pi_i  d_{\text{CMMD}}(\bar{D_t}, D_s^i) \\
        &\leq d_{\text{CMMD}}(D_t, \bar{D_t}) + \sup_{i, j \in \mathcal{H}} d_{\text{CMMD}}(D_s^i, D_s^j).
    \end{aligned}
    \label{eq:retracted CMMD}
\end{equation}
Moreover, we also investigate the relationship between the MMD and CMMD distances based on Definitions $1$ and $2$:
\begin{equation}
    \begin{aligned}
        d_{\text{MMD}}(&D_s^i, D_t)=||\int_x k(x, \cdot)d(\phi(D_s^i) - \phi(D_t))||_{\mathcal{H}_k} \\
        &= ||\int_x k(x, \cdot)d(\frac{1}{n}\sum_{i=1}^{n}(\phi_i(D_s^i) - \phi_i(D_t)))||_{\mathcal{H}_k} \\
        &\leq ||\int_x k(x, \cdot) \sum_{i=1}^{n} \sup_{\phi_i \in \Phi_i} d(\phi_i(D_s^i) - \phi_i(D_t))||_{\mathcal{H}_k} \\
        & = d_{\text{CMMD}}(D_s^i, D_t).
    \end{aligned}
\end{equation}

Based on the above preparations, we now derive the generalization error bound of the model on the unseen target domain.
Recalling that Lemma~$1$ indicates the generalization error bound between two different distributions. Considering the pair of the $i$-th source domain and the target domain, the following holds with the probability of at least $1-\delta$:
\begin{equation}
    \begin{aligned}
        \mathcal{R}_t[h] \leq & \mathcal{R}_s^i[h] + d_{\text{CMMD}}(D_t, D_s^i) + \frac{2}{m} (
        E_{x \sim D_s^i}[\sqrt{tr(K_{D_s^i})}] \\
        & + E_{x\sim D_t}[\sqrt{tr(K_{D_t})}]) + 
        2\frac{\log(\frac{2}{\sigma})}{2m} + \epsilon. \\
    \end{aligned}
\end{equation}
We then generalize the above inequality to the multi-source scenario, where the ideal target domain could be expressed as a weighted combination of different source domains. 
We weight the generalization error of each source-target pair with $\pi$ where $\sum_{i=1}^K \pi_i = 1$ and calculate their sum:
\begin{equation}
    \begin{aligned}
        \mathcal{R}_t[h] \leq &  \sum_{i=1}^{N}\pi_i \mathcal{R}_s^i[h] +  \sum_{i=1}^{N}\pi_i d_{\text{CMMD}}(D_t, D_s^i) \\
        &+ \frac{2}{m} (
            \sum_{i=1}^{N}\pi_i E_{x \sim D_s^i}[\sqrt{tr(K_{D_s^i})}] \\
        & + E_{x\sim D_t}[\sqrt{tr(K_{D_t})}]) + 
        2\frac{\log(\frac{2}{\sigma})}{2m} + \epsilon.
    \end{aligned}
    \label{eq:weighted sum}
\end{equation}
By replacing the CMMD distance in Eq.~(\ref{eq:weighted sum}) with the retracted CMMD distance in Eq.~(\ref{eq:retracted CMMD}), we arrive at Theorem~1.

\section{Additional Experiments}

We conduct additional experiments to verify the effectiveness of our DomainDrop, including: 1) The effects of different inserted positions of DomainDrop in the network; 2) The experiments on the DomainBed benchmark.

\textbf{Different inserted positions of DomainDrop.} 
We here investigate where to insert DomainDrop in the network. 
Given a standard ResNet with four residual blocks, we train different models by taking different blocks as candidates and randomly selecting a block to activate DomainDrop at each iteration.
The results are reported in Tab.~\ref{table: Layers}. 
The first line represents the results of the baseline model, which is trained using all source domains directly on the ResNet-$18$ (\ie, DeepAll \cite{zhou2020deep}). 
We observe that no matter where DomainDrop is inserted, the model consistently outperforms the baseline model by a significant margin, \eg, $3.15\%$ ($84.15\%$ vs. $81.00\%$) with DomainDrop in Block~1.
The results indicate that our DomainDrop is effective in enhancing the robustness of channels to domain shifts at different network layers.
Furthermore, we find that inserting DomainDrop into all blocks of the network leads to the highest performance, exceeding the baseline model by $5.66\%$ ($86.66\%$ vs. $81.00\%$), indicating that suppressing domain-sensitive channels in all training stages will result in the best generalization ability. 
Based on the analysis, we insert DomainDrop into all network blocks in our all experiments.

\begin{table}[tb!] 
    \begin{center}
    \caption{
        Effect (\%) on different inserted posotions of DomainDrop. 
        B$1-4$ represent four residual blocks of the ResNet architecture.
        The experiment is conducted on PACS dataset with ResNet-$18$ backbone. The best performance is marked as \textbf{bold}.
        }
    \vspace{-0.1cm}
    \label{table: Layers}
    \resizebox{\linewidth}{!}{
    \begin{tabular}{cccc | cccc |c}
        \hline
        \multicolumn{4}{c|}{Position} & \multicolumn{5}{c}{PACS} \\
        \hline
        B1 & B2 & B3 & B4 & Art & Cartoon & Photo & Sketch & Avg.\\
        \hline
        - & - & - & - & 80.31{\tiny $\pm$ 1.54} & 76.65{\tiny $\pm$ 0.48} & 95.38{\tiny $\pm$ 0.12} & 71.67{\tiny $\pm$ 1.49} & 81.00 \\
        \hline
        \checkmark & - & - & - & 81.10{\tiny $\pm$ 0.76} & 78.88{\tiny $\pm$ 0.69} & 94.72{\tiny $\pm$ 0.45} & 81.92{\tiny $\pm$ 0.69} & 84.15 \\
        - & \checkmark & - & - & 80.71{\tiny $\pm$ 0.71} & 79.25{\tiny $\pm$ 0.44} & 94.85{\tiny $\pm$ 0.35} & 82.16{\tiny $\pm$ 1.35} & 84.24 \\
        - & - & \checkmark & - & 82.52{\tiny $\pm$ 0.72} & 79.44{\tiny $\pm$ 0.46} & 95.76{\tiny $\pm$ 0.16} & 79.35{\tiny $\pm$ 1.17} & 84.27 \\
        - & - & - & \checkmark & 81.15{\tiny $\pm$ 0.98} & 78.58{\tiny $\pm$ 0.81} & 95.39{\tiny $\pm$ 0.40} & 79.74{\tiny $\pm$ 1.47} & 83.72 \\
        \hline
        \checkmark & \checkmark & - & - & 81.15{\tiny $\pm$ 1.03} & 79.44{\tiny $\pm$ 0.30} & 95.99{\tiny $\pm$ 0.49} & 83.13{\tiny $\pm$ 0.48} & 84.93 \\
        \checkmark & \checkmark & \checkmark & - & 83.84{\tiny $\pm$ 0.70} & 80.02{\tiny $\pm$ 0.37} & 96.29{\tiny $\pm$ 0.23}  & 83.23{\tiny $\pm$ 0.53} & 85.87 \\
        \rowcolor{mygray} \checkmark & \checkmark & \checkmark & \checkmark &  \textbf{84.47}{\tiny $\pm$ 0.77} & \textbf{80.50}{\tiny $\pm$ 0.56} & \textbf{96.83}{\tiny $\pm$ 0.21} & \textbf{84.83}{\tiny $\pm$ 0.67} & \textbf{86.66} \\
        \hline
    \end{tabular}}
    \end{center}
    \vspace{-0.2cm}
\end{table}

\begin{table}[tb!]
    \centering
    \caption{Performance (\%) comparisons with the start-of-the-art DG approaches on the DomainBed benchmark. We compare with $12$ DG algorithms on the following five multi-domain datasets: VLCS \cite{torralba2011unbiased}, PACS \cite{li2017deeper}, OfficeHome \cite{venkateswara2017deep}, TerraInc \cite{beery2018recognition}, and DomainNet \cite{peng2019moment}. The network architecture is ResNet-$50$. We use the validation set from source domains for the model selection.}
    \vspace{0.3cm}
    \resizebox{1.0\linewidth}{!}{
    \begin{tabular}{l|c|ccccc|c}
      \hline
      \textbf{Method} & Venue & VLCS & PACS & OfficeHome & TerraInc & DomainNet & Avg. \\
      \hline
      ERM \cite{gulrajani2020search} & ICLR'20 & 77.5 $\pm$ 0.4 & 85.5 $\pm$ 0.2 & 66.5 $\pm$ 0.3 & 46.1 $\pm$ 1.8 & 40.9 $\pm$ 0.1 & 63.3 \\
      RSC \cite{huang2020self} & ECCV'20 & 77.1 $\pm$ 0.5 & 85.2 $\pm$ 0.9 & 65.5 $\pm$ 0.9 & 46.6 $\pm$ 1.0 & 38.9 $\pm$ 0.5 & 62.7 \\
      SagNet \cite{nam2021reducing} & CVPR'21 & 77.8 $\pm$ 0.5 & 86.3 $\pm$ 0.2 & 68.1 $\pm$ 0.1 & 48.6 $\pm$ 1.0 & 40.3 $\pm$ 0.1 & 64.2 \\
      SelfReg \cite{kim2021selfreg} & ICCV'21 & 77.5 $\pm$ 0.0 & 86.5 $\pm$ 0.3 & 69.4 $\pm$ 0.2 & 51.0 $\pm$ 0.4 & 44.6 $\pm$ 0.1 & 65.8 \\
      FISH \cite{shi2021gradient} & ICLR'21 & 77.8 $\pm$ 0.3 & 85.5 $\pm$ 0.3 & 68.6 $\pm$ 0.4 & 45.1 $\pm$ 1.3 & 42.7 $\pm$ 0.2  & 63.9 \\
      W2D \cite{huang2022two} & CVPR'22 & - & 83.4 $\pm$ 0.3 & 63.5 $\pm$ 0.1 & 44.5 $\pm$ 0.5 & - & - \\
      XDED \cite{lee2022cross} & ECCV'22 & 74.8 $\pm$0.0 & 83.8 $\pm$ 0.0 & 65.0 $\pm$ 0.0 & 42.5 $\pm$ 0.0 & - & - \\
      GVRT \cite{min2022grounding} & ECCV'22 & 79.0 $\pm$ 0.2 & 85.1 $\pm$ 0.3 &  70.1 $\pm$ 0.1 & 48.0 $\pm$ 0.2 & 44.1 $\pm$ 0.1 & 65.2 \\
      MIRO \cite{cha2022domain} & ECCV'22 & 79.0 $\pm$ 0.0 & 85.4 $\pm$ 0.4 & \textbf{70.5} $\pm$ 0.4 & 50.4 $\pm$ 1.1 & 44.3 $\pm$ 0.2 & 65.9 \\
      PTE \cite{min2022grounding} & ECCV'22 & 79.0 $\pm$ 0.2 & 85.1 $\pm$ 0.3 & 70.1 $\pm$ 0.1 & 48.0 $\pm$ 0.2 & 44.1 $\pm$ 0.1 & 65.2 \\
      EQRM \cite{eastwood2022probable} & NeurIPS'22 & 77.8 $\pm$ 0.6 & 86.5 $\pm$ 0.2 & 67.5 $\pm$  0.1 & 47.8 $\pm$ 0.6 & 41.0 $\pm$ 0.3 & 64.1 \\
      DAC-SC \cite{lee2023decompose} & CVPR'23 & 78.7 $\pm$ 0.3 & 87.5 $\pm$ 0.1 & 70.3 $\pm$ 0.2 & 44.9 $\pm$ 0.1 & \textbf{46.5} $\pm$ 0.3 & 65.6 \\
      \hline 
      \rowcolor{mygray} DomainDrop & Ours & \textbf{79.8} $\pm$ 0.3  & \textbf{87.9} $\pm$ 0.3 & 68.7 $\pm$ 0.1 & \textbf{51.5} $\pm$ 0.4 & 44.4 $\pm$ 0.5 & \textbf{66.5} \\
      \hline
    \end{tabular}}
    \label{tab:domainbed}
    \vspace{-0.2cm}
  \end{table}

\textbf{Experiments on DomainBed.} 
We conducted experiments on the DomainBed benchmark \cite{gulrajani2020search}, including VLCS, PACS, OfficeHome, TerraInc, and DomainNet. 
The network is trained using Adam optimizer for $5000$ iterations with a learning rate of $5e-5$ and batch size of $64$. 
The experiments are repeated three times, and the averaged accuracy is reported in Tab.~\ref{tab:domainbed}.
We observe that our DomainDrop can consistently achieve better performance than ERM (a strong baseline in DomainBed) on all datasets, \eg, outperforming ERM by $2.4\%$ ($87.9\%$ vs. $85.5\%$) on PACS and $5.4\%$ ($51.5\%$ vs. $46.1\%$) on TerraInc.
The experimental results demonstrate the effectiveness of our method on various DG benchmark datasets.
Moreover, DomainDrop obtained the highest average accuracy among all the compared methods, exceeding the SOTA method DAC-SC \cite{lee2023decompose} by $0.9\%$ ($66.5\%$ vs. $65.6\%$), indicating that our method can significantly improve the model generalization ability.

\section{Analytical Experiments}

We conduct experiments to analyze the effectiveness of our method, including: 1) We discuss why tackle the DG issue on feature channels; 2) We quantify the channel robustness to domain shifts in each network layer; 3) We measure the domain gap of feature maps extracted by the model; 4) We provide visual explanations of our DomainDrop.

\begin{figure}[tb!]
    \begin{center}
    \includegraphics[width=0.8\linewidth]{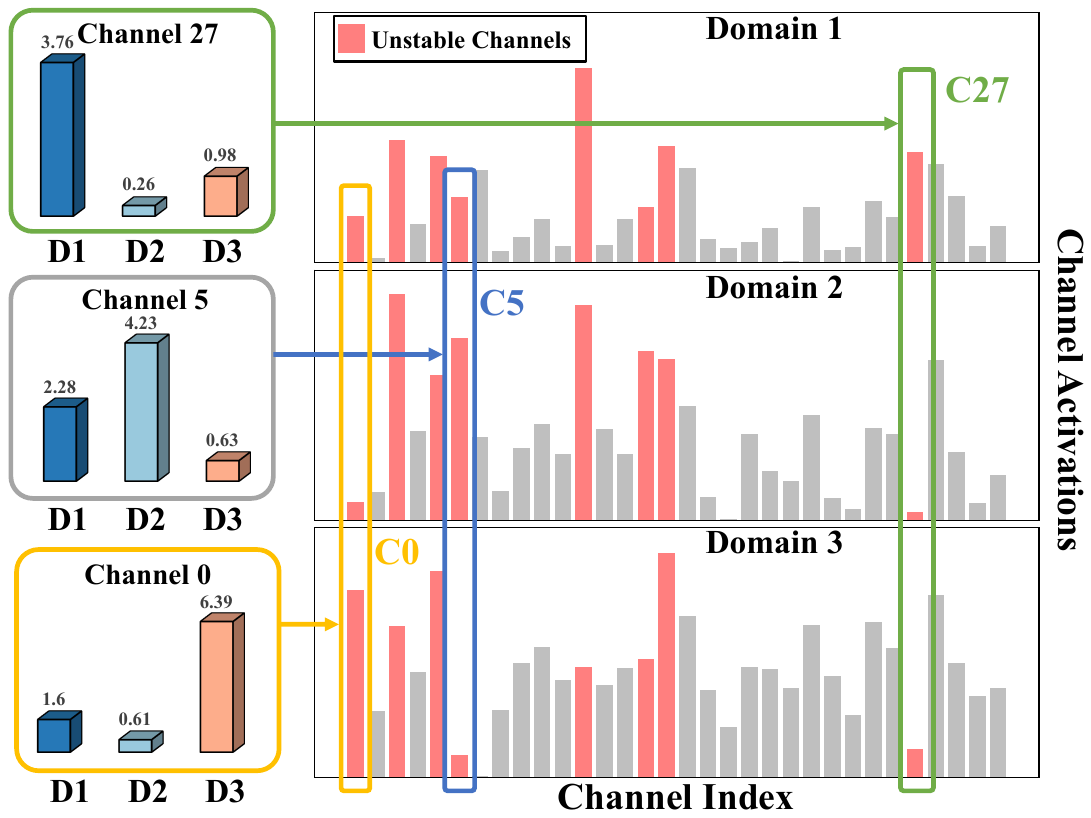}
    \end{center}
    \vspace{-0.2cm}
    \caption{
        The activation value of the channels $1-32$ in the last block of ResNet-$18$ across different domains. The experiment is conducted on the PACS dataset with Art as the target domain. 
    }
    \label{fig:channel activation}
\end{figure}

\textbf{\textit{Why tackle DG on feature channels.}}
Different from traditional DG methods that constrain the entire network, recent methods have focused on learning domain-invariant features in middle layers via domain augmentations \cite{wang2022feature,zhou2021domain} or local penalizations \cite{shi2020informative,wang2019learning}. 
However, recent work \cite{ding2022domain} has indicated that these methods typically perturb or penalize specific pre-defined features, \eg, style statistics \cite{zhou2021domain} or local textures \cite{shi2020informative}, which could neglect other domain-specific features and affect model generalization. 
In this paper, we propose to analyze the DG issue from a novel perspective of channel robustness to domain shifts. 
\textit{Our key insight is that if a channel captures domain-invariant patterns, its activations should remain stable across different domains.}
As shown in Fig.~\ref{fig:channel activation}, we observe that numerous 
channels exhibit limited robustness to domain shifts (\ie, the red bars).
The findings motivate us to focus on enhancing channel robustness to domain shifts.

\textbf{Channel robustness to domain shifts.} 
To enhance the generalization ability of the models to the unseen target domain, we wish the model to learn general and comprehensive domain-invariant features from source domains.
Ideally, we hope each channel of the representations is activated by category-related information while being invariant across domains, making the whole representation sufficient for classification. 
Inspired by previous work \cite{wang2022feature}, we exploit the averaged activation for each class in each domain to estimate the robustness of each channel to domain shifts.
Specifically, for the $i$-th channel in the $l$-th middle layer, we first calculate its averaged activation in the $k$-the domain:
\begin{equation}
    a_i^l = \frac{1}{n_k} \sum_{j=1}^{n_k} GAP(F_l(x_j))_i,
\end{equation}
where $F_l(\cdot)$ is the feature maps in the $l$-th middle layer and $GAP(\cdot)$ denotes the global average pooling layer.
Then we compute the standard deviation of the $i$-th channel activation among different domains.
We present the results in Tab.~\ref{tab:channel sensitivity}.
We observe that compared with Baseline, RSC \cite{huang2020self} and I$^{2}$-Drop \cite{shi2020informative} present lower channel sensitivity to domain shifts in the last layer (\ie, Block.~$4$) since they can regularize the model to learning domain-invariant features. 
However, since these methods are only suitable for specific layers (\ie, RSC for the deepest layer and I$^{2}$-Drop for the shallowest layer), they cannot adequately counter the overfitting issue.
The SOTA DG methods MixStyle \cite{zhou2021domain} can increase the feature diversity at multiple layers, but it does not explicitly remove domain-specific features, thus failing to reduce channel sensitivity adequately. 
In contrast, the lowest standard deviation that DomainDrop achieves indicates that our method can learn more domain-invariant representations, showing the superiority of our framework.

\begin{table}[tb!]
    \centering
    \caption{
        The standard deviation of channel activations for samples from different domains. 
        Block.~1$-$4 represent four residual blocks of the ResNet architecture.
        The lower the standard deviation, the more robust the channel is to domain shifts.
    }
    \vspace{0.3cm}
    \resizebox{1.0\linewidth}{!}{
        \setlength{\tabcolsep}{4.5mm}{
    \begin{tabular}{l|cccc}
    \hline
    Sensitivity & Block. 1 & Block. 2 & Block. 3 & Block. 4 \\
    \hline
    Baseline & 4.43 & 2.06 & 1.54 & 7.84 \\
    RSC \cite{huang2020self} & 4.22 & 1.94 & 1.63 & 7.23 \\
    I$^2$-Drop \cite{shi2020informative} & 4.03 & 1.89 & 1.58 & 7.42 \\
    MixStyle \cite{zhou2021domain} & 4.30 & 2.03 & 1.51 & 7.16 \\
    FACT \cite{xu2021fourier} & 4.83 & 2.07 & 1.57 & 7.52 \\
    \hline
    \rowcolor{mygray} DomainDrop (Ours) & \textbf{3.85} & \textbf{1.56} & \textbf{1.04} & \textbf{5.94} \\
      \hline
    \end{tabular}
    }}
    \label{tab:channel sensitivity}
  \end{table}

\textbf{Domain gap of extracted features maps.} 
To investigate the influence of our framework, we also calculate the inter-domain distance (across all source domains) of the feature maps extracted by the model on various datasets, including PACS, OfficeHome, and VLCS.
Following previous DG method \cite{wang2022feature}, we calcute the inter-domain gap as:
\begin{equation}
    d=\frac{2}{K(K-1)}\sum_{k_1=1}^{K}\sum_{k_2=k_1+1}^{K} ||\overline{F}_{k_1} - \overline{F}_{k_2}||_2,
\end{equation}
where $K$ is the number of source domains, $\overline{F}_{k_1}$ and $\overline{F}_{k_2}$ denote the averaged feature maps of all samples from the $k_1$-th and $k_2$-th domain, respectively. 
As shown in Tab.~\ref{tab:domain gap}, we can observe that compared to the baseline, DomainDrop can effectively narrow the inter-domain gap among source domains on all datasets, indicating that our method can suppress domain-specific features and encourage the model to learn domain-invariant features during training.


\begin{table}[tb!]
    \centering
    \caption{The inter-domain distribution gap ($\times 100$) of the extracted features by our method.
    For PACS, we take Art Painting as the target domain and the others as all source domains.
    For OfficeHome, the target domain is Real-World and the others are source domains. 
    For VLCS, we adopt Sun as the target domain and the others as source domains.
    The smaller the inter-domain distance, the better the generalization performance of the model. 
    }
    \vspace{0.2cm}
    \resizebox{\linewidth}{!}{
        \setlength{\tabcolsep}{7mm}{
    \begin{tabular}{l|cccc}
      \hline
    Method & PACS & OfficeHome & VLCS \\
    \hline
    Baseline & 17.57 & 11.56 & 16.65 \\
    \rowcolor{mygray} DomainDrop (Ours) & \textbf{11.82} & \textbf{8.58} & \textbf{14.21} \\
      \hline
    \end{tabular}
    }}
    \label{tab:domain gap}
  \end{table}

\begin{figure}[tb!]
    \begin{center}
    \includegraphics[width=\linewidth]{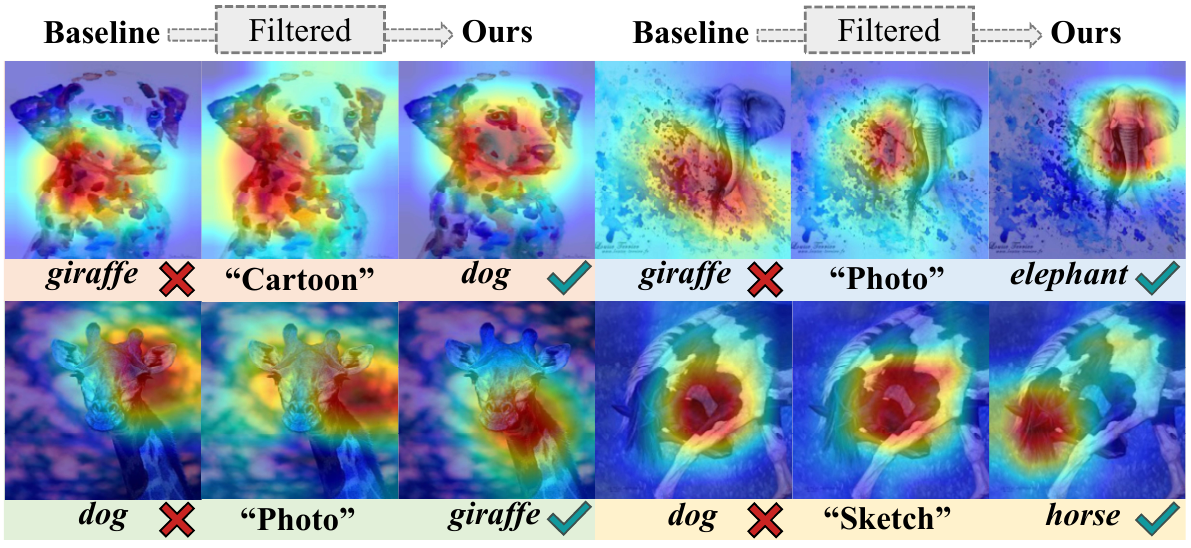}
    \end{center}
    \caption{
        Visualization of attention maps of the last convolutional layer on PACS with Art Painting as the target domain. The backbone used in the experiment is ResNet-$18$. 
        For each sample, the first column is the category attention map of baseline, the middle column is the domain attention map generated by domain discriminator, and the last column is the attention map of DomainDrop.
    }
    \label{fig:CAM Drop}
    \vspace{-0.1cm}
\end{figure}

\textbf{Visual explanations.}
To provide visual evidence of the effectiveness of DomainDrop in reducing domain-specific features, we utilized GradCAM \cite{selvaraju2017grad} to generate attention maps of the last conventional layer for both the baseline (DeepAll) and DomainDrop models. 
The results are presented in Fig.~\ref{fig:CAM Drop}. 
As we can see, the baseline model captures a considerable amount of domain-specific information, as indicated by the overlap between the category attention map (column $1$) and the domain attention map (column $2$). 
On the other hand, DomainDrop can discard domain-specific features while retaining domain-invariant features, leading to more generalized attention maps that focus on representative information for object classification (column $3$). 
For instance, in the case of the dog image, the model needs to focus on the dog's face as one of the representative features to classify, which is precisely captured by DomainDrop. 
In contrast, the baseline focuses on spot texture features, which results in misclassification. 
These results suggest that DomainDrop can effectively reduce the sensitivity of the model to domain shifts and learn more generalized features, making it a promising method for DG tasks.


\maketitle
\ificcvfinal\thispagestyle{empty}\fi

\end{document}